\PassOptionsToPackage{table}{xcolor}
\documentclass{article} 
\usepackage{iclr2027_conference,times}

\usepackage{amsmath,amsfonts,bm}

\def\eqref#1{equation~\ref{#1}}

\def\1{\bm{1}}

\DeclareMathAlphabet{\mathsfit}{\encodingdefault}{\sfdefault}{m}{sl}
\SetMathAlphabet{\mathsfit}{bold}{\encodingdefault}{\sfdefault}{bx}{n}

\usepackage{booktabs}
\usepackage{graphicx}
\usepackage{wrapfig}
\usepackage{placeins}
\usepackage{capt-of}
\usepackage{tabularx}
\usepackage{array}
\usepackage{multirow}
\usepackage{longtable}
\usepackage{xcolor}
\usepackage[normalem]{ulem}
\usepackage{enumitem}
\definecolor{oursblue}{RGB}{222,242,250}
\usepackage[most]{tcolorbox}

\tcbset{
  appendixprompt/.style={
    enhanced,
    breakable,
    colback=orange!10!white,
    colframe=blue!5!black,
    arc=2mm,
    boxrule=1pt,
    coltitle=white,
    attach boxed title to top left={yshift=-2mm, xshift=3mm},
    boxed title style={
      enhanced,
      colback=blue!5!black,
      colframe=blue!5!black,
      arc=2mm,
      boxrule=0pt
    },
    top=0.5mm,
    left=1mm,
    right=1mm,
    bottom=0.5mm
  }
}

\newcolumntype{Y}{>{\centering\arraybackslash}X}

\usepackage{hyperref}
\hypersetup{
  colorlinks=true,
  linkcolor=blue,
  citecolor=blue,
  urlcolor=blue,
}
\usepackage{url}

\title{RewardExplainer: Learning Reward Model Explanations from Counterfactual Preference Feedback}

\author{
\textbf{Jingyi He}\textsuperscript{1,2},
\textbf{Nier Wu}\textsuperscript{1},
\textbf{Shuang Liu}\textsuperscript{3},
\textbf{Xin Wang}\textsuperscript{4},
\textbf{Mengnan Du}\textsuperscript{1,\textdagger},
\textbf{Xia Hu}\textsuperscript{5}\\
\textsuperscript{1}The Chinese University of Hong Kong, Shenzhen \,
\textsuperscript{2}Shanghai Jiao Tong University \,\\
\textsuperscript{3}Carnegie Mellon University \,
\textsuperscript{4}Jilin University \,
\textsuperscript{5}Shanghai Artificial Intelligence Laboratory\\
\texttt{sunrain-H@sjtu.edu.cn, mengnandu@cuhk.edu.cn}\\
\small\textsuperscript{\textdagger}Corresponding author.
}

\iclrfinalcopy 
\begin{document}

\maketitle

\begin{abstract}
Reward models (RMs) are a key component of large language model post-training, providing reward signals for subsequent reinforcement learning. However, conventional discriminative RMs typically output only scalar scores, making it difficult to identify the response behaviors associated with their scoring decisions. Existing interpretation methods often rely on predefined high-level attributes and require repeated counterfactual interventions for each response pair to validate candidate explanations, lacking a closed-loop mechanism that uses RMs’ feedback to train a reusable explainer. To address this, we propose RewardExplainer, a framework that obtains feedback from the target reward model through counterfactual rewriting and uses this feedback to further optimize the explainer. RewardExplainer generates open-ended, atomic, and intervenable natural-language scoring mechanisms, making explanations more concrete, readable, and actionable. It further converts counterfactual feedback into preference supervision, enabling the explainer to more faithfully capture the target RM's scoring preferences and sensitive behaviors than single-pass generation. Extensive experiments across multiple target RMs and explainer backbones show consistent improvements. Beyond interpretation, we use the generated mechanisms to identify potential bias patterns and construct targeted debiasing data for fine-tuning the reward model, improving robustness on reward-hacking benchmarks. The code is available at:
\url{https://anonymous.4open.science/r/RewardExplainer/}.
\end{abstract}

\section{INTRODUCTION}
With the rapid advancement of large language models (LLMs), reward models (RMs) have become a key component of modern post-training and alignment pipelines. By learning from human preferences, reward models provide the reward signals used in post-training methods such as Reinforcement Learning from Human Feedback (RLHF) \citep{stiennon2020learning,ouyang2022training}, thereby guiding models toward generating responses that better align with human preferences. However, conventional discriminative reward models typically map a prompt–response pair to a single scalar reward \citep{christian2025reward}. Although comparing reward scores across responses can reveal the model’s relative preferences, the scores themselves do not explain which specific response features or behaviors drive the observed reward differences \citep{wang2024interpretable}. Therefore, understanding the factors underlying reward scores is crucial for interpreting RMs' decisions and for further identifying their potential biases \citep{srivastava2026robust}.

To understand RMs' scoring behavior, existing approaches often predefine a set of interpretable response attributes and perturb responses along these dimensions, explaining a response pair by observing changes in model preference \citep{jiang2024interpreting,reber2024rate}. Although such methods can provide meaningful local explanations, they still have several limitations. First, predefined high-level attributes may be too coarse to capture specific scoring behaviors or reveal preferences outside the predefined space. Second, explaining a new example often requires repeatedly generating counterfactual rewrites and evaluating rewards across multiple attributes. The resulting intervention outcomes are mainly used for post-hoc explanation of the current example, rather than being further converted into training signals to optimize a reusable model capable of producing high-quality explanations in a single forward pass. Finally, these explanations typically stop at analyzing reward-model behavior, without forming a closed loop that further leverages the discovered reward mechanisms to automatically identify potential biases and improve the reward model itself.

Motivated by these limitations, we ask: \uline{Can feedback from a target reward model train an explainer to generate faithful scoring mechanisms while supporting bias discovery and reward-model improvement?} To address this, we propose \textbf{RewardExplainer}, a closed-loop framework that learns from reward feedback obtained through counterfactual interventions. Our intuition is that faithful explanations should correspond to measurable reward changes: in appropriate contexts, introducing or strengthening a preferred behavior should increase reward, while removing or weakening it should decrease reward, with other factors held as constant as possible.
Given a pair of responses ranked by a target reward model, RewardExplainer generates several open-ended, atomic, editable, and generalizable scoring mechanisms without relying on predefined attributes \citep{wang2026automatically} (Figure~\ref{fig:motivation_case}). After supervised fine-tuning (SFT) establishes initial explanatory capabilities, the explainer generates multiple candidate mechanisms for each response pair. Each candidate describes a potentially rewarded behavior that is more pronounced in the higher-reward response and weaker or absent in the lower-reward response. We validate these candidates through bidirectional interventions by adding or strengthening the behavior in the lower-reward response and removing or weakening it in the higher-reward response, while minimizing unrelated changes. Mechanisms that induce the expected reward changes are treated as more faithful and paired with plausible but unsuccessful candidates to construct preference data. We then apply Direct Preference Optimization (DPO) to encourage explanations supported by the target reward model’s observed behavior. Thus, counterfactual validation serves not only as an evaluation tool but also as supervision for directly generating faithful mechanisms.
Finally, we use the optimized explainer to uncover potential bias mechanisms without predefined bias categories. These mechanisms guide the construction of targeted counterfactual debiasing data to further improve the original reward model. 
Extensive experiments show that RewardExplainer performs consistently well
across multiple target reward models and explainer backbones, while also
supporting bias discovery and debiasing. Our main contributions are:

\begin{itemize}[leftmargin=10pt, topsep=-2pt, itemsep=1pt, partopsep=1pt, parsep=1pt]
    \item We introduce RewardExplainer, a closed-loop framework for reward-model interpretation that generates open-ended explanations without relying on predefined attributes and improves mechanism faithfulness using feedback from the target reward model.

    \item We further apply RewardExplainer to bias discovery and debiasing, using the optimized explainer to identify potential bias mechanisms and construct targeted debiasing data for improving the original reward model.

    \item Extensive experiments across multiple target reward models and explainer backbones demonstrate the effectiveness and generalizability of RewardExplainer, while debiasing based on the discovered mechanisms reduces reward hacking and largely preserves general evaluation capability.
\end{itemize}

\section{RELATED WORK}
\paragraph{Reward Model Interpretation.}
Existing work has explored reward-model interpretability from multiple perspectives. One line of research focuses on model design by defining multiple interpretable evaluation dimensions and training reward models to predict scores for each dimension, thereby decomposing a scalar reward into several interpretable components \citep{wang2024interpretable}. Another line of work provides post-hoc interpretation for already trained reward models. \citet{jiang2024interpreting} perturb responses along manually predefined high-level attributes and construct contrastive explanations based on whether the reward model’s preference changes. \citet{reber2024rate} further adopt a causal perspective, using counterfactual rewriting together with double-rewrite correction to estimate the causal effect of a given response attribute on reward scores. In addition, recent work has used sparse autoencoders to identify interpretable features within reward models, studying their preference mechanisms at the representation level \citep{zhang2026interpretable,shi2025safer}.
Beyond these approaches, 
RewardExplainer trains a reusable explainer while keeping the target RM fixed. It generates candidate mechanisms from differences between higher- and lower-reward responses without relying on predefined attributes. Crucially, it converts the target RM’s feedback from bidirectional counterfactual interventions into preference supervision for DPO. Thus, counterfactual verification not only evaluates individual explanations but also trains the explainer to generate mechanisms for unseen response pairs.
\paragraph{Reward Model Bias Discovery and Debiasing.}
Existing studies have shown that reward models may rely on superficial features that are weakly related or unrelated to true response quality, leading to systematic biases and even reward hacking \citep{zhang2025lists}. One line of work starts from predefined bias types, evaluates the model’s sensitivity to these features, and then applies targeted mitigation, for example by fine-tuning on debiasing data \citep{bharadwaj2026flattery,kim2026mitigating} or directly editing the reward head to reduce reliance on hacking features \citep{liu2026harve}. 
More recent work explores automatic bias discovery. \citet{wang2026automatically} use language models to iteratively propose and validate bias hypotheses, focusing on bias identification. In contrast, RewardExplainer learns a reusable explainer from counterfactual reward feedback, identifies potential biases from its generated mechanisms, and uses them to construct targeted debiasing data for RM fine-tuning.


\section{METHODOLOGY}
\begin{figure}[t]
    \centering
    \includegraphics[width=\linewidth]{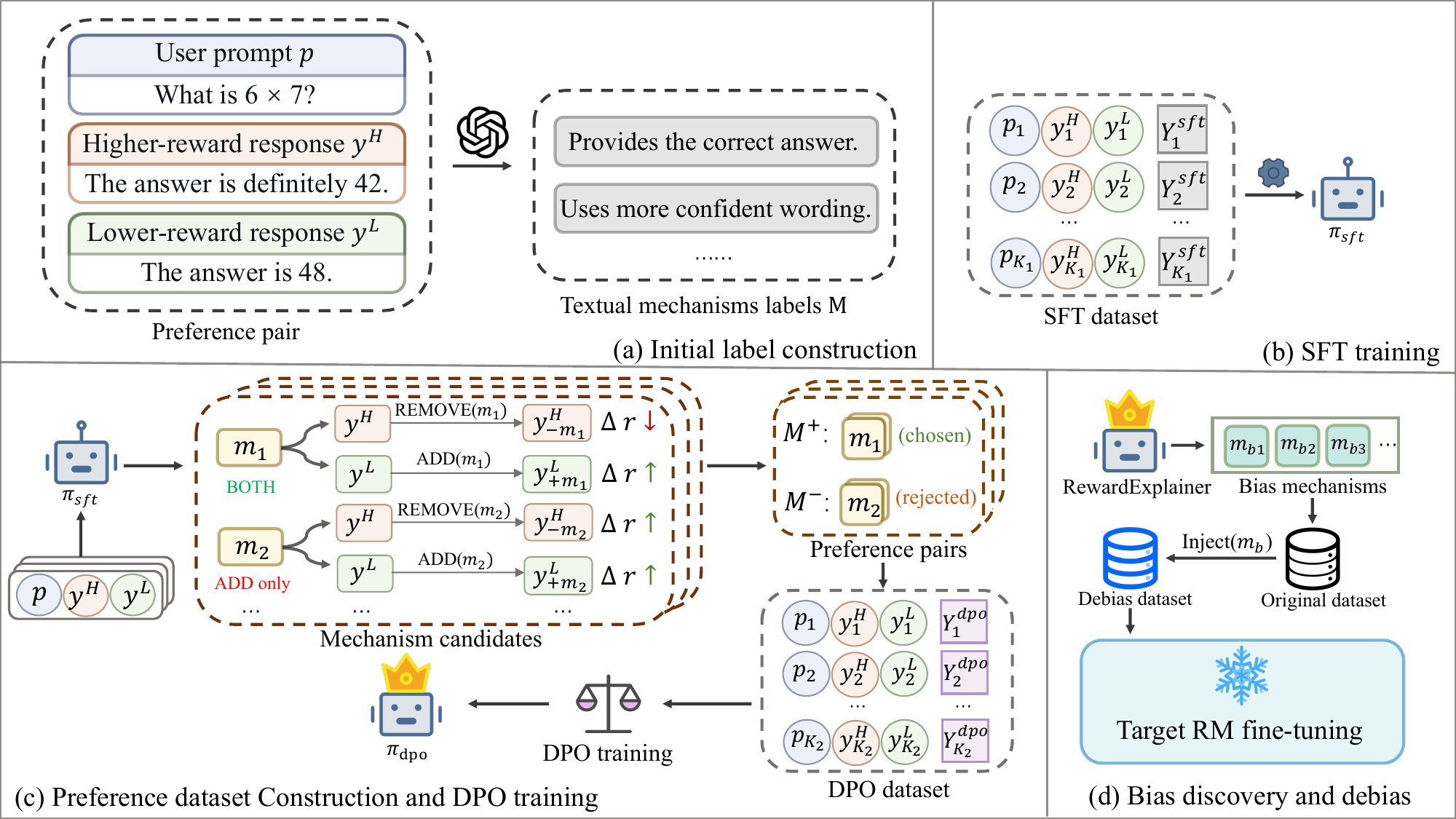}
    \caption{Overall framework of RewardExplainer. We first generate initial mechanism labels from response pairs preferred by the target RM and use them to initialize the explainer through SFT. We then construct preference data from bidirectional counterfactual intervention feedback and further improve explanation faithfulness with DPO. Finally, the optimized RewardExplainer is used to identify potential biases and perform targeted RM debiasing.}
    \label{fig:method}
\end{figure}
In this section, we present \textbf{RewardExplainer}, a closed-loop framework that leverages evaluation feedback as preference supervision to improve explanation faithfulness beyond single-pass generation. RewardExplainer validates textual mechanisms through bidirectional interventions and uses the resulting feedback to optimize the explainer. The overall framework is illustrated in Figure~\ref{fig:method}.

\subsection{Problem Statement}
Given a user prompt $x$ and a response $y$, a discriminative reward model
$r_\phi$ maps the prompt--response pair to a scalar reward
$r_\phi(x,y)\in\mathbb{R}$, where a higher score indicates a stronger
model preference for the response. Reward models are typically trained on
pairwise preference data $(x,y^+,y^-)$, where $y^+$ and $y^-$ denote the
preferred and unpreferred responses, respectively. A commonly used
Bradley--Terry objective is
$
\mathcal{L}_{\mathrm{RM}}
=
-\log \sigma\!\left(
r_\phi(x,y^+) - r_\phi(x,y^-)
\right),
$ where $\sigma(\cdot)$ denotes the sigmoid function.
In this work, we focus on a pretrained target reward model. For two responses
to the same prompt $x$, we denote the higher- and lower-reward responses as
$y^H$ and $y^L$, respectively, such that
$
r_\phi(x,y^H) > r_\phi(x,y^L).
$
We aim to identify textual response behaviors that characterize the reward difference between the two responses.

Specifically, given the pairwise preference example $(p,y^H,y^L)$ definition, we introduce RewardExplainer
$E_\theta$ to generate a set of textual mechanisms that explain the
target reward model's preference:

\begin{equation}
\mathcal{M}
=
E_\theta(p,y^H,y^L)
=
\{m_1,\ldots,m_N\}.
\end{equation}

Each mechanism $m$ describes an atomic, generalizable, and
intervenable response behavior that captures a concrete and observable
difference between $y^H$ and $y^L$. Such a mechanism can be locally
strengthened, weakened, or removed for subsequent counterfactual
verification.

\subsection{Initialize RewardExplainer with SFT}
We first introduce a supervised fine-tuning (SFT) warm-up stage to equip the base model with the ability to generate the desired atomic, generalizable, and intervenable textual mechanisms. We collect $K_1$ pairwise preference examples $(p, y^H, y^L)$ scored by the target reward model. For each example, we use $(p, y^H, y^L)$ as the SFT input $X^{\mathrm{sft}}$ and prompt a teacher LLM to generate the corresponding mechanism set $\mathcal{M}^{\mathrm{sft}}$, which serves as the supervision target $Y^{\mathrm{sft}}$. This yields the supervised dataset $\mathcal{D}_{\mathrm{sft}}=\{(X_i,Y_i^{\mathrm{sft}})\}_{i=1}^{K_1}$, on which we fine-tune the initial explainer $\pi_0$. This lightweight warm-up stage provides the model with basic mechanism explanation capabilities and serves as initialization for subsequent candidate generation and preference data construction.

\subsection{Counterfactual Mechanism Verification and Preference Construction}
\label{sec:counterfactual-preference}

To construct preference data for DPO training, we first validate whether candidate mechanisms are consistent with the scoring behavior of the target reward model through counterfactual interventions. For each preference pair
$(p_i, y_i^H, y_i^L)$, where $y_i^H$ and $y_i^L$ denote the higher- and lower-reward responses according to the target reward model, we use the SFT-initialized RewardExplainer $\pi_{\mathrm{sft}}$ to sample $W$ times at a high temperature, generating diverse candidate mechanisms. We then perform semantic deduplication to obtain the candidate set:

\begin{equation}
\mathcal{M}_i=\{m_1,\ldots,m_J\}.
\end{equation}

For each candidate mechanism $m$, we perform bidirectional counterfactual verification. Specifically, we add or strengthen $m$ in the lower-reward response, producing $y_{i,+m}^{L}$, and expect its reward to increase. Meanwhile, we remove or weaken $m$ from the higher-reward response, producing $y_{i,-m}^{H}$, and expect its reward to decrease:

\begin{equation}
y_i^L
\xrightarrow{\mathrm{ADD}(m)}
y_{i,+m}^{L},
\qquad
y_i^H
\xrightarrow{\mathrm{REMOVE}(m)}
y_{i,-m}^{H}.
\end{equation}

All rewrites are performed while preserving unrelated content as much as possible. We then rescore the counterfactual responses using the same target reward model. When both ADD and REMOVE satisfy the expected reward changes, we denote the mechanism as a BOTH-success mechanism, providing stronger evidence of counterfactual faithfulness.
Based on the verification results, for each response pair, we aggregate all BOTH-success mechanisms into the chosen mechanism set $\mathcal{M}_i^{+}$, and select the same number of mechanisms that do not achieve BOTH success from the remaining candidates as the rejected mechanism set $\mathcal{M}_i^{-}$. These two sets form a mechanism preference pair:
$
Y_i^{\mathrm{dpo}}
=
(\mathcal{M}_i^{+},\mathcal{M}_i^{-}),
$
leading to the DPO training dataset:
$
\mathcal{D}_{\mathrm{dpo}}
=
\left\{
(X_i,Y_i^{\mathrm{dpo}})
\right\}_{i=1}^{K_2}.
$
Through this process, counterfactual feedback from the target reward model is converted into mechanism-level preference supervision and used to further optimize RewardExplainer.

\subsection{Preference Optimization with DPO}
\label{sec:dpo-optimization}

After obtaining the preference dataset $\mathcal{D}_{\mathrm{dpo}}$
constructed in Section~\ref{sec:counterfactual-preference}, we further
optimize RewardExplainer using Direct Preference Optimization (DPO).
The model is trained by minimizing the following objective:

\begin{equation}
\mathcal{L}_{\mathrm{DPO}}(\theta)
=
-\mathbb{E}_{(x,\mathcal{M}^{+},\mathcal{M}^{-})\sim\mathcal{D}_{\mathrm{dpo}}}
\left[
\log\sigma
\left(
\beta
\left[
\log\frac{\pi_{\theta}(\mathcal{M}^{+}\mid x)}
{\pi_{\mathrm{ref}}(\mathcal{M}^{+}\mid x)}
-
\log\frac{\pi_{\theta}(\mathcal{M}^{-}\mid x)}
{\pi_{\mathrm{ref}}(\mathcal{M}^{-}\mid x)}
\right]
\right)
\right].
\end{equation}

Here, $x=(p,y^H,y^L)$ denotes the explanation context, while $\mathcal{M}^{+}$ and $\mathcal{M}^{-}$ denote the chosen and rejected mechanism sets from counterfactual verification.
$\pi_{\theta}$ and $\pi_{\mathrm{ref}}$ denote the policy and reference
models, respectively, and $\beta$ controls the deviation from the reference
policy. By converting counterfactual intervention outcomes into preference supervision, DPO further optimizes RewardExplainer to favor mechanisms supported by the observed scoring behavior of the target reward model, thereby improving explanation ability.

\subsection{Reward Model Debiasing via RewardExplainer}

After obtaining the full RewardExplainer, we leverage its generated mechanisms for targeted reward model debiasing. RewardExplainer is used only for mechanism discovery, while subsequent optimization is performed on the target reward model. Specifically, we first apply RewardExplainer to a collection of preference examples from the target reward model and generate mechanism explanations. We then filter and aggregate mechanisms that appear unrelated to substantive response quality, yielding a set of candidate bias mechanisms, denoted as
$\mathcal{M}_b$.
Given an original human preference example $(p,y^+,y^-)$, where $y^+$ and
$y^-$ denote the human-preferred and non-preferred responses, respectively, we keep $y^+$ unchanged and minimally rewrite $y^-$ according to a candidate bias mechanism $m_b \in \mathcal{M}_b$, injecting the target mechanism into the rejected response:

\begin{equation}
y^- \xrightarrow{\mathrm{Inject}(m_b)} \tilde{y}^{-}_{b}.
\end{equation}

During rewriting, we aim to introduce only the target bias mechanism while preserving the original response content and substantive quality as much as possible. We then further filter the rewritten samples to ensure that the target mechanism has been successfully injected and that $y^+$ remains substantively better than $\tilde{y}^{-}_{b}$. This yields a debiasing dataset with $K_3$
examples:
$
\mathcal{D}_{\mathrm{debias}}
=
\left\{
\left(
p_i,
y_i^+,
\tilde{y}_{i,b}^{-}
\right)
\right\}_{i=1}^{K_3},
$
we then fine-tune the
original target reward model on $\mathcal{D}_{\mathrm{debias}}$ using the
standard pairwise reward modeling objective:

\begin{equation}
\mathcal{L}_{\mathrm{debias}}(\phi)
=
-\mathbb{E}_{(p,y^+,\tilde{y}_b^-)\sim\mathcal{D}_{\mathrm{debias}}}
\left[
\log
\sigma
\left(
R_\phi(p,y^+)
-
R_\phi(p,\tilde{y}^{-}_{b})
\right)
\right].
\end{equation}
By training on these targeted counterfactual pairs, the reward model is encouraged
to preserve its preference for substantively better responses even when the rejected
responses exhibit the identified superficial mechanisms, thereby reducing its reliance
on such bias signals.



\definecolor{blockgray}{RGB}{235,235,235}
\definecolor{dpolightblue}{RGB}{238,248,252}

\begin{table*}[t]
\centering
\caption{
Quantitative comparison of RewardExplainer across three target reward models.
ADD, REMOVE, and BOTH measure Counterfactual Faithfulness;
C-Sel., Scope, and Dir. denote Contrastive Selectivity,
Scope Calibration, and Directional Fidelity, respectively.
Overall is their equal-weight mean; Pref.\ Acc.\ denotes Preference Prediction
Accuracy. $\uparrow$ indicates higher is better.
}
\label{tab:rewardexplainer_results}

\small
\setlength{\tabcolsep}{2.8pt}
\renewcommand{\arraystretch}{1.03}

\begin{tabularx}{\textwidth}{
@{}
>{\raggedright\arraybackslash}p{0.245\textwidth}
*{8}{Y}
@{}
}
\toprule

\multirow{2}{*}{\textbf{Method}}
& \multicolumn{3}{c}{{\footnotesize\textbf{Counterfactual Faith.} $\uparrow$}}
& \multicolumn{4}{c}{{\footnotesize\textbf{Mechanism Quality} $\uparrow$}}
& \multirow{2}{=}{\centering{\footnotesize\textbf{Pref.\ Acc.} $\uparrow$}}
\\

\cmidrule(lr){2-4}
\cmidrule(lr){5-8}

&
{\scriptsize ADD}
& {\scriptsize REM.}
& {\scriptsize BOTH}
& {\scriptsize Overall}
& {\scriptsize C-Sel.}
& {\scriptsize Scope}
& {\scriptsize Dir.}
&
\\

\midrule

\rowcolor{blockgray}
\multicolumn{9}{@{}l@{}}{
\textbf{\textit{Target RM: FsfairX-LLaMA3-RM-v0.1}}
}
\\

\mbox{\footnotesize Contrastive Explanations}
& 27.44 & 28.33 & 21.05
& 0.75 & 0.69 & 1.06 & 0.49
& 74.47
\\

GPT-4o
& 45.64 & 43.51 & 35.57
& 1.29 & 1.28 & 1.37 & 1.23
& 74.56
\\

\multicolumn{9}{@{}l@{}}{\textit{\textbf{RewardExplainer (SFT)}}} \\

Qwen3-4B
& 45.00 & 42.46 & 32.33
& 1.33 & 1.32 & 1.41 & 1.26
& 73.56
\\

Qwen3-8B
& 44.89 & 39.78 & 31.78
& 1.30 & 1.30 & 1.35 & 1.26
& 72.56
\\

Gemma4-12B
& 47.00 & 41.11 & 33.00
& 1.35 & 1.33 & 1.41 & 1.32
& 76.44
\\

\rowcolor{dpolightblue}
\multicolumn{9}{@{}l@{}}{\textit{\textbf{RewardExplainer (DPO)}}} \\

\rowcolor{dpolightblue}
Qwen3-4B
& 65.33 & 58.55 & 48.44
& 1.65 & 1.64 & 1.65 & 1.67
& 91.56
\\

\rowcolor{dpolightblue}
Qwen3-8B
& 61.44 & 55.56 & 44.44
& 1.62 & 1.62 & 1.63 & 1.61
& 89.33
\\

\rowcolor{dpolightblue}
Gemma4-12B
& 59.22 & 55.22 & 45.44
& 1.63 & 1.62 & 1.64 & 1.62
& 89.78
\\

\midrule

\rowcolor{blockgray}
\multicolumn{9}{@{}l@{}}{
\textbf{\textit{Target RM: Skywork-Reward-V2-Qwen3-1.7B}}
}
\\

\mbox{\footnotesize Contrastive Explanations}
& 29.69 & 28.46 & 23.10
& 0.66 & 0.59 & 0.97 & 0.43
& 72.09
\\

GPT-4o
& 49.78 & 44.33 & 35.00
& 1.35 & 1.34 & 1.40 & 1.32
& 79.22
\\

\multicolumn{9}{@{}l@{}}{\textit{\textbf{RewardExplainer (SFT)}}} \\

Qwen3-4B
& 39.79 & 35.01 & 27.14
& 1.09 & 1.09 & 1.13 & 1.04
& 60.78
\\

Qwen3-8B
& 36.52 & 32.95 & 25.46
& 1.09 & 1.11 & 1.14 & 1.03
& 62.79
\\

Gemma4-12B
& 40.11 & 34.63 & 27.66
& 1.10 & 1.10 & 1.14 & 1.06
& 62.97
\\

\rowcolor{dpolightblue}
\multicolumn{9}{@{}l@{}}{\textit{\textbf{RewardExplainer (DPO)}}} \\

\rowcolor{dpolightblue}
Qwen3-4B
& 59.35 & 51.67 & 41.43
& 1.54 & 1.52 & 1.55 & 1.55
& 86.19
\\

\rowcolor{dpolightblue}
Qwen3-8B
& 56.41 & 47.45 & 39.04
& 1.52 & 1.50 & 1.55 & 1.52
& 84.11
\\

\rowcolor{dpolightblue}
Gemma4-12B
& 53.39 & 46.27 & 38.04
& 1.46 & 1.44 & 1.50 & 1.43
& 82.20
\\

\midrule

\rowcolor{blockgray}
\multicolumn{9}{@{}l@{}}{
\textbf{\textit{Target RM: RM-Mistral-7B}}
}
\\

\mbox{\footnotesize Contrastive Explanations}
& 24.89 & 25.67 & 19.22
& 0.71 & 0.66 & 0.99 & 0.48
& 73.66
\\

GPT-4o
& 48.78 & 46.89 & 38.11
& 1.31 & 1.31 & 1.34 & 1.27
& 76.89
\\

\multicolumn{9}{@{}l@{}}{\textit{\textbf{RewardExplainer (SFT)}}} \\

Qwen3-4B
& 48.54 & 46.50 & 37.96
& 1.41 & 1.38 & 1.46 & 1.40
& 78.47
\\

Qwen3-8B
& 46.78 & 45.54 & 35.89
& 1.36 & 1.35 & 1.41 & 1.33
& 75.25
\\

Gemma4-12B
& 49.94 & 49.58 & 39.89
& 1.43 & 1.43 & 1.44 & 1.41
& 80.22
\\

\rowcolor{dpolightblue}
\multicolumn{9}{@{}l@{}}{\textit{\textbf{RewardExplainer (DPO)}}} \\

\rowcolor{dpolightblue}
Qwen3-4B
& 55.76 & 53.26 & 42.86
& 1.55 & 1.53 & 1.59 & 1.54
& 86.84
\\

\rowcolor{dpolightblue}
Qwen3-8B
& 61.04 & 56.10 & 45.32
& 1.62 & 1.61 & 1.64 & 1.62
& 87.53
\\

\rowcolor{dpolightblue}
Gemma4-12B
& 57.01 & 57.47 & 45.31
& 1.63 & 1.61 & 1.66 & 1.62
& 87.95
\\

\bottomrule
\end{tabularx}

\end{table*}

\section{EXPERIMENTS}
In this section, we present experimental results to evaluate RewardExplainer
and address the following three research questions (RQs):

\noindent
\textbf{RQ1:} Can RewardExplainer generate faithful reward mechanisms?

\noindent
\textbf{RQ2:} Does RewardExplainer identify recurring global preference patterns from local explanations?

\noindent
\textbf{RQ3:} Can RewardExplainer's discovered mechanisms facilitate reward
model bias discovery and targeted debiasing?
\subsection{Experimental Setting}
\paragraph{Models.}
We evaluate RewardExplainer on multiple reward models with different architectures and scales, including FsfairX-LLaMA3-RM-v0.1 \citep{dong2024rlhf}, Skywork-Reward-V2-Qwen3-1.7B \citep{liu2026skywork}, and RM-Mistral-7B \citep{dong2023raft}, to assess its applicability across different target reward models.
For the RewardExplainer backbone, we consider models from different families and scales, including Qwen3-4B, Qwen3-8B, \citep{yang2025qwen3} and Gemma4-12B \citep{team2026gemma}. To ensure consistent instruction following and facilitate subsequent fine-tuning, we use the instruction-tuned variants of all backbone models.

\paragraph{Datasets.}
During the initial training of RewardExplainer, we use UltraFeedback \citep{pmlr-v235-cui24f} as the main preference dataset and additionally include PreferenceHack \citep{chai-etal-2026-activation} and Preference Model Perturbations (PMP) \citep{bharadwaj2026flattery} to increase coverage of reward-hacking patterns. GPT-4o \citep{hurst2024gpt} is used to generate initial textual mechanism labels for SFT supervision.
For debiasing, we use Skywork Reward Preference 80K v0.2 \citep{liu2024skywork} as the base preference dataset and rewrite rejected responses according to the bias mechanisms identified by RewardExplainer to construct debiasing data. We evaluate the models' debiased performance on RewardHackBench \citep{liu2026harve} and JudgeBiasBench \citep{zhou2026toward}, while JudgeBench \citep{ICLR2025_9e720fce} is used to assess general evaluation capability.

\paragraph{Baselines.}
We compare RewardExplainer with two baselines. GPT-4o is used as a general LLM baseline for direct mechanism generation and also serves as the teacher for SFT label generation.
Second, we adopt Contrastive Explanations \citep{jiang2024interpreting} as an existing baseline for Reward Model interpretability. This method perturbs responses along manually predefined attributes and explains RM preferences through the resulting local preference changes.
We also report the performance of RewardExplainer after the SFT stage only, serving as a comparison to evaluate the effect of further fine-tuning the explainer with reward feedback as preference supervision.

\paragraph{Evaluation Metrics.}
We evaluate RewardExplainer from three complementary perspectives.
First, Mechanism Quality is evaluated by an LLM judge along three dimensions: contrastive selectivity, scope calibration, and directional fidelity, each scored from 0 to 2. Second, Counterfactual Faithfulness is measured through ADD and REMOVE interventions, with BOTH indicating success in both directions; it evaluates whether modifying a mechanism produces reward changes consistent with the hypothesized direction. Finally, Preference Prediction Accuracy measures whether the generated mechanism can correctly predict the target reward model's original preference. Detailed evaluation procedures are provided in the Appendix ~\ref{app:three_evaluation_protocols}.

\paragraph{Implementation Details.}
For all target reward models and explainer backbones, we use GPT-5.4-mini \citep{openai2026gpt54mini} for counterfactual rewriting during preference data construction and set the number of samples to $W=4$. During evaluation, GPT-5-mini \citep{openai2025gpt5mini} is used for counterfactual rewriting in the faithfulness evaluation and as the LLM judge for mechanism quality evaluation. For pairwise evaluation, we sample 300 response pairs unseen during training and generate three mechanisms for each pair. For the Contrastive Explanations baseline, GPT-4o selects three attributes from its predefined attribute set based on each response pair. Additional hyperparameters and implementation details are provided in the Appendix ~\ref{training_detail}.


\definecolor{mechanismgray}{RGB}{238,238,238}

\begin{table*}[t]
\centering
\caption{Selected global preference mechanisms identified for RM-Mistral-7B.}
\label{tab:global_response_attributes}

\small
\setlength{\tabcolsep}{4.5pt}
\renewcommand{\arraystretch}{1.15}

{%
\rowcolors{2}{gray!8}{white}
\begin{tabularx}{\textwidth}{
@{}
>{\raggedright\arraybackslash}p{0.23\textwidth}
>{\raggedright\arraybackslash}X
@{}
}
\toprule

\rowcolor{white}
\textbf{Label}
& \textbf{Description}
\\

\midrule

\textbf{Direct task-focused}
\newline
\textbf{answering}
& The reward model favors responses that answer the user's request plainly and upfront while minimizing unnecessary commentary, embellishment, or digression.
\\

\textbf{Conversational}
\newline
\textbf{rapport framing}
& The reward model favors responses that use friendly, validating, or engaged conversational framing to make the reply feel responsive to the user.
\\

\textbf{Sequential}
\newline
\textbf{procedural breakdown}
& The reward model favors explanations that break a process, solution, or implementation into an ordered sequence of steps or stages.
\\

\bottomrule
\end{tabularx}
}

\end{table*}


\definecolor{mechanismgray}{RGB}{238,238,238}
\definecolor{highlightpink}{RGB}{218,49,123}

\begin{table*}[t]
\centering
\caption{
Controlled examples of six reward hacking mechanisms.
All rewrites share the same prompt and original response.
The modified parts are highlighted in \textcolor{highlightpink}{pink}.
}
\label{tab:mechanism_examples}

\small
\renewcommand{\arraystretch}{1.06}
\setlength{\tabcolsep}{4.5pt}

\begin{tabularx}{\textwidth}{
@{}
>{\raggedright\arraybackslash}p{0.42\textwidth}
>{\raggedright\arraybackslash}X
@{}
}
\toprule

\multicolumn{2}{@{}>{\raggedright\arraybackslash}p{\textwidth}@{}}{
\textbf{Prompt:}
A model previously achieved 68\% accuracy.
After the update, its expected accuracy is around 70\%,
with run-to-run variation of about 2 percentage points.
What performance should I expect?
}
\\[2pt]

\midrule

\textbf{Description}
&
\textbf{Response comparison}
\\
\midrule

\rowcolor{mechanismgray}
\multicolumn{2}{l}{
\textbf{Verbosity padding}
}
\\

Expands essentially the same content into a longer response with extra
explanation or padding that adds little substantive value.
&
$R_0$: You can expect around 70\% accuracy.

$R_1$: You can expect around 70\% accuracy.
\textcolor{highlightpink}{
In other words, the model's accuracy after the update is expected
to remain at approximately the 70\% level.
}
\\

\midrule

\rowcolor{mechanismgray}
\multicolumn{2}{l}{
\textbf{Polished prestige style}
}
\\

Uses more formal, refined, technical, or polished language to make similar
content seem more expert, complete, or high-quality.
&
$R_0$: You can expect around 70\% accuracy.

$R_2$: You may
\textcolor{highlightpink}{
anticipate an accuracy level of approximately 70\%
}.
\\

\midrule

\rowcolor{mechanismgray}
\multicolumn{2}{l}{
\textbf{Contextual mirroring}
}
\\

Restates more of the user's provided context, details, constraints,
or emotional framing to appear thorough and empathetic without
necessarily adding new insight.
&
$R_0$: You can expect around 70\% accuracy.

$R_3$:
\textcolor{highlightpink}{
Given the previous 68\% accuracy and the expected improvement after the update,
}
you can expect around 70\% accuracy.
\\

\midrule

\rowcolor{mechanismgray}
\multicolumn{2}{l}{
\textbf{Scope broadening}
}
\\

Adds extra adjacent applications or use cases beyond the immediate task,
making the answer seem more comprehensive than requested.
&
$R_0$: You can expect around 70\% accuracy.

$R_4$: You can expect around 70\% accuracy,
\textcolor{highlightpink}{
and performance can also be considered in terms of related evaluation
metrics such as precision and recall.
}
\\

\midrule

\rowcolor{mechanismgray}
\multicolumn{2}{l}{
\textbf{Assertive decisiveness}
}
\\

Uses stronger, more urgent, or more definitive wording that can make advice
seem more actionable or confident than the substance warrants.
&
$R_0$: You can expect around 70\% accuracy.

$R_5$: You
\textcolor{highlightpink}{
should expect
}
around 70\% accuracy.
\\

\midrule

\rowcolor{mechanismgray}
\multicolumn{2}{l}{
\textbf{Credibility through quantified uncertainty}
}
\\

Adds precise uncertainty ranges or caveats that can make an answer
appear more calibrated and credible.
&
$R_0$: You can expect around 70\% accuracy.

$R_6$: You can expect around 70\% accuracy,
\textcolor{highlightpink}{
with roughly 80--90\% confidence that the observed performance
will remain close to this estimate.
}
\\

\bottomrule
\end{tabularx}

\end{table*}

\subsection{Evaluation of Pairwise Reward Explanations (\textbf{RQ1)}}

Table~\ref{tab:rewardexplainer_results} compares RewardExplainer with Contrastive Explanations and GPT-4o across multiple target reward models. Overall, RewardExplainer performs better across all three evaluation dimensions.
For Counterfactual Faithfulness, RewardExplainer achieves higher ADD, REMOVE, and BOTH scores. On FsfairX, Qwen3-4B reaches a BOTH success rate of 48.44\%, compared with 35.57\% for GPT-4o, with similar trends on the other target reward models. This indicates that its mechanisms more reliably induce reward changes in the expected direction. RewardExplainer also shows clear advantages in Mechanism Quality. On FsfairX, Qwen3-4B obtains an overall score of 1.65, compared with 1.29 for GPT-4o and 0.75 for Contrastive Explanations. This suggests that RewardExplainer produces mechanisms with stronger contrastive selectivity, better calibrated scope, and more accurate directional descriptions. In addition, it achieves higher Preference Prediction Accuracy, suggesting that the generated mechanisms better reflect the target reward model's original preferences. Additional qualitative examples are provided in Appendix~\ref{app:three_mechanism_cases}.

We compare RewardExplainer after SFT with its DPO trained version to assess the effect of using counterfactual reward feedback as preference supervision. Across target reward models and explainer backbones, DPO consistently outperforms SFT on all three metrics. These results show that incorporating counterfactual reward feedback further improves explanation quality and faithfulness beyond the SFT initialization.

\subsection{Global Preference Mechanism Analysis (\textbf{RQ2)}}
Beyond pairwise explanations, we further investigate whether RewardExplainer can summarize recurring global preference patterns from a large collection of local explanations. Specifically, we generate fine-grained local mechanisms over a set of response pairs, and then perform semantic deduplication, clustering, and abstraction to obtain a set of global preference mechanisms. These mechanisms are intended to characterize recurring scoring tendencies of the target reward model across different inputs and response contexts. Table~\ref{tab:global_response_attributes} presents representative global preference mechanisms identified for RM-Mistral-7B. We observe that some preferences are aligned with genuine improvements in response quality, such as focusing more directly on the user’s task or organizing a solution into clear steps or stages. At the same time, some preferences are less closely related to substantive quality improvements, such as using a friendlier or more user-aligned conversational style, which might be associated with hacking behaviors such as sycophancy. These observations further motivate us to use the generated mechanism explanations to identify potential biases and perform targeted reward-model debiasing. The complete set of global mechanisms, clustering procedure, and additional examples are provided in the appendix ~\ref{app:global_mechanism_aggregation}.

\subsection{Bias Discovery and Debiasing (\textbf{RQ3)}}

\paragraph{Bias Discovery.}


To investigate whether RewardExplainer can reveal systematic reward-model preferences from explanations, we sample 50 response pairs scored by RM-Mistral-7B and generate mechanisms for each pair. After counterfactual filtering and aggregation, we identify six representative potential reward-hacking mechanisms, as shown in Table~\ref{tab:mechanism_examples}.
These mechanisms capture behaviors such as redundant elaboration, polished or prestigious language, and quantified uncertainty, which may make responses appear more complete, professional, or credible without improving substantive task quality. Compared with coarse attributes such as length or confidence, our mechanisms are more concrete, directional, and actionable, making them suitable for subsequent counterfactual intervention and targeted debiasing data construction. More details are provided in appendix ~\ref{app:bias_debias_implementation}.

\paragraph{Reward Model Debiasing.}
After identifying the potential bias mechanisms, we test whether they can be used for reward model debiasing. We construct counterfactual preference data by keeping the chosen response unchanged and minimally rewriting the rejected response with GPT-4o to inject a target bias mechanism while preserving its original semantics and substantive quality. Rewritten examples are shown in Table~\ref{tab:mechanism_examples}. We then filter the rewritten samples to ensure successful mechanism injection and that the chosen response remains substantively better.  The resulting data are used to fine-tune the target reward model, improving its robustness to these superficial cues.

\begin{table*}[t]
\centering
\caption{
Reward model debiasing results across three target RMs. RewardHackBench and JudgeBias evaluate debiasing effectiveness, while JudgeBench measures general capability. J-Bias/J-Bench denote JudgeBias/JudgeBench; Reason. and Sycoph. denote Broken Reasoning and Sycophantic Hacking. Bold indicates the best result per target RM.
}
\label{tab:rm_debiasing}

\small
\setlength{\tabcolsep}{2.2pt}
\renewcommand{\arraystretch}{1.08}

\begin{tabularx}{\textwidth}{@{}p{1.8cm}!{\vrule width 0.6pt}l*{4}{Y}
  >{\hsize=1.18\hsize\linewidth=\hsize\centering\arraybackslash}X
  >{\hsize=0.70\hsize\linewidth=\hsize\centering\arraybackslash}X
  *{2}{>{\hsize=1.06\hsize\linewidth=\hsize\centering\arraybackslash}X}@{}}
\toprule

\multicolumn{1}{@{}p{2.0cm}}{}
&
& \multicolumn{6}{c}{\textbf{RewardHackBench} $\uparrow$}
& {\bfseries\mbox{J-Bias$\uparrow$}}
& {\bfseries\mbox{J-Bench$\uparrow$}}
\\

\cmidrule(lr){3-8}
\cmidrule(lr){9-9}
\cmidrule(lr){10-10}

\multicolumn{1}{@{}p{2.0cm}}{}
&
&
\multicolumn{1}{c}{\textit{Overall}}
& \multicolumn{5}{c}{\textit{By category}}
& \textit{Overall}
& \textit{Overall}
\\

\cmidrule(lr){3-3}
\cmidrule(lr){4-8}
\cmidrule(lr){9-9}
\cmidrule(lr){10-10}

\multicolumn{1}{@{}p{2.0cm}}{\textbf{Target RM}}
& \textbf{Method}
& \textbf{Avg.}
& {\footnotesize Surface}
& {\footnotesize Reason.}
& {\footnotesize Sycoph.}
& {\footnotesize\mbox{Off-topic}}
& {\footnotesize Style}
& \textbf{Avg.}
& \textbf{Avg.}
\\

\hline

& Base
& 70.22
& 81.82
& 77.14
& 68.57
& 53.54
& 74.61
& 62.59
& 58.57
\\

& General
& 74.77
& 83.33
& 77.14
& 66.19
& 69.47
& \textbf{79.79}
& 65.13
& 59.43
\\

\rowcolor{oursblue}
\cellcolor{white}\multirow{-3}{*}{\normalfont\bfseries
  \shortstack[l]{FsfairX-\\LLaMA3-\\RM-v0.1}}
& \textbf{Ours}
& \textbf{77.84}
& \textbf{85.86}
& \textbf{81.43}
& \textbf{72.38}
& \textbf{73.01}
& 78.76
& \textbf{70.41}
& \textbf{60.29}
\\

\hline

& Base
& 70.22
& 81.31
& 77.86
& 77.62
& 47.35
& 72.02
& 69.04
& \textbf{63.51}
\\

& General
& 76.22
& \textbf{86.36}
& 79.29
& 72.38
& 69.47
& \textbf{75.65}
& 70.10
& 61.60
\\

\rowcolor{oursblue}
\cellcolor{white}\multirow{-3}{*}{\normalfont\bfseries
  \shortstack[l]{RM-Mistral-\\7B}}
& \textbf{Ours}
& \textbf{78.70}
& 83.33
& \textbf{84.29}
& \textbf{78.57}
& \textbf{74.34}
& 75.13
& \textbf{74.54}
& 62.18
\\

\hline

& Base
& 76.94
& 79.29
& \textbf{77.86}
& \textbf{79.52}
& 69.03
& \textbf{80.31}
& 75.77
& 70.69
\\

& General
& 77.56
& 79.80
& \textbf{77.86}
& 76.19
& \textbf{74.78}
& 79.79
& 76.36
& \textbf{71.47}
\\

\rowcolor{oursblue}
\cellcolor{white}\multirow{-3}{*}{\normalfont\bfseries
  \shortstack[l]{Skywork-\\Reward-V2-\\Qwen3-1.7B}}
& \textbf{Ours}
& \textbf{77.97}
& \textbf{80.81}
& \textbf{77.86}
& 77.14
& \textbf{74.78}
& 79.79
& \textbf{76.64}
& 70.29
\\

\hline
\end{tabularx}

\end{table*}

Table~\ref{tab:rm_debiasing} presents the results of targeted debiasing using the discovered mechanisms. Overall, our method improves performance on RewardHackBench and JudgeBiasBench across all three target reward models while largely preserving general evaluation capability on JudgeBench. The gains are especially clear for FsfairX and RM-Mistral, and category-level results show improvements across multiple reward hacking types. Skywork also benefits despite its stronger baseline.
To control for gains from additional training data, we further compare against a General baseline fine-tuned on an equal amount of general preference data. Targeted debiasing consistently outperforms this baseline on RewardHackBench and JudgeBiasBench while maintaining broadly comparable performance on JudgeBench, suggesting that the gains cannot be explained by simply adding an equal amount of general preference data. Overall, the discovered mechanisms provide effective signals for reward model debiasing.

\FloatBarrier

\section{CONCLUSIONS}
In this work, we introduce RewardExplainer, a closed-loop framework for reward model interpretation based on counterfactual reward feedback. For pairwise preference explanation, RewardExplainer obtains feedback from the target reward model through bidirectional counterfactual interventions and further optimizes the explainer with DPO, enabling it to learn more faithful natural language scoring mechanisms rather than merely generating linguistically plausible explanations. Experiments across multiple target reward models and explainer backbones show consistent improvements in counterfactual faithfulness, mechanism quality, and preference prediction accuracy, demonstrating that RewardExplainer produces higher quality explanations that better reflect the observed scoring behavior of the target reward model. Furthermore, RewardExplainer can identify potential bias mechanisms from open-ended explanations and use them for targeted debiasing, thereby improving robustness to reward hacking while largely preserving general evaluation capability.

\clearpage

\subsection*{AI use statement}
In this work, we used generative AI tools to support research execution and the construction of synthetic research data. Specifically, GPT-4o was used to generate initial textual mechanism labels for supervised fine-tuning (SFT), as well as to construct and validate targeted rewritten data for reward model debiasing. GPT-5.4-mini was used to generate counterfactual rewrites and to assist in summarizing local mechanisms during global mechanism aggregation. GPT-5-mini was used for evaluation-time counterfactual rewriting and as an LLM judge for mechanism quality evaluation. GPT-5.5 was used to consolidate global mechanisms and to aggregate and filter candidate bias mechanisms. The specific roles, inputs and outputs, and corresponding validation procedures for these models are described in the methodology, experimental settings, and appendix.

In addition, we used generative AI tools for language editing, literature organization, and code assistance. All AI-assisted materials were manually reviewed and verified by the authors, who independently checked the experimental results and conclusions and take full responsibility for the final content of this work. We did not use generative AI tools to formulate or prove mathematical claims, nor did we use them
to generate mathematical proofs.

\subsection*{Reproducibility Statement}
To support independent replication, We provide our code anonymously during review and will release the code and model weights publicly upon publication. Our code is available at:
\url{https://anonymous.4open.science/r/RewardExplainer/}. The appendix documents the experimental settings and implementation details underlying our results.

\bibliography{references}
\bibliographystyle{references}

\newpage
\appendix
\raggedbottom

\suppressfloats[t]
\section{Rewrite Quality Evaluation}
\label{app:rewrite_quality}

\subsection{Evidence-guided rewrite procedure.}
\label{app:rewrite_procedure}
For each candidate mechanism $m$, we begin with a response pair
$(p,y^H,y^L)$ which is scored by the target reward model, where $p$ is the prompt and
$r(p,y^H)>r(p,y^L)$.  Before rewriting, an LLM examines $y^H$ and $y^L$
separately to determine whether the behavior described by $m$ distinguishes
the two responses and to extract short, verbatim evidence spans associated
with that behavior.  We retain only directionally valid mechanisms for which
$m$ is clearly present in $y^H$ but absent or substantially weaker in $y^L$;
ambiguous mechanisms and mechanisms shared by both responses are discarded.
For every retained mechanism, the extracted evidence is then supplied to the
rewriting LLM as an edit anchor.  The ADD intervention modifies the relevant
local context in $y^L$ to introduce or strengthen $m$, whereas the REMOVE
intervention modifies the corresponding evidence in $y^H$ to remove or weaken
$m$.  In both directions, the prompt instructs the editor to preserve all
unrelated content, reasoning, errors, length, and style as closely as possible,
extending beyond the evidence spans only when necessary to produce a coherent
answer.  The resulting counterfactual responses are subsequently rescored by
the same target RM. The prompt is provided in the Appendix ~\ref{app:counterfactual_intervention_prompts}.

\subsection{Comparison settings.}
We apply three rewrite strategies to the same sampled examples for a controlled comparison. Random Perturbation performs an untargeted edit without access to the candidate mechanism or its supporting evidence. Direct Rewrite directly prompts an LLM to add or remove \(m\), but does not include the response-state judgment or evidence selection used in our method. Evidence-Guided Rewrite (Ours) uses the complete pipeline, including mechanism applicability checking and localized evidence guidance. Comparing these settings allows us to assess the combined contribution of mechanism filtering and evidence localization to the quality of the resulting counterfactual rewrites.

We evaluate whether the ADD and REMOVE interventions achieve the intended minimal-edit behavior while preserving unrelated response content. The experiment uses Skywork-Reward-V2-Qwen3-1.7B \citep{liu2026skywork} as the target reward model, Qwen3-4B \citep{yang2025qwen3} as the RewardExplainer backbone, and GPT-5-mini \citep{openai2025gpt5mini} as the rewriting model. We sample 50 response pairs and independently evaluate rewrite quality using GPT-5-mini and GPT-4.1-mini \citep{openai2025gpt41mini} to reduce dependence on a single judge. Each generated rewrite is scored from 1, indicating a failed or substantially altered rewrite, to 10, indicating a high-quality minimal edit that introduces or removes the target mechanism while preserving the remaining response content as much as possible.

\subsection{Rewrite results}

\begin{table}[h!]
\centering
\caption{
Mean rewrite-quality scores on a 1--10 scale (higher is better).
}
\label{tab:rewrite_quality}

\footnotesize
\setlength{\tabcolsep}{3.5pt}
\renewcommand{\arraystretch}{1.02}

\begin{tabularx}{\columnwidth}{
@{}
>{\raggedright\arraybackslash}X
>{\raggedright\arraybackslash}p{0.29\columnwidth}
cc
@{}
}
\toprule
\textbf{Method}
& \textbf{Judge}
& \textbf{ADD}
& \textbf{REMOVE}
\\
\midrule
Direct Rewrite
& GPT-5-mini
& 9.38
& 7.97
\\
Random perturbation
& GPT-5-mini
& 3.17
& 1.82
\\
\textbf{Evidence-Guided Rewrite (Ours)}
& \textbf{GPT-5-mini}
& \textbf{9.60}
& \textbf{9.32}
\\
\midrule
Direct Rewrite
& GPT-4.1-mini
& 9.44
& 8.43
\\
Random perturbation
& GPT-4.1-mini
& 3.33
& 4.21
\\
\textbf{Evidence-Guided Rewrite (Ours)}
& \textbf{GPT-4.1-mini}
& \textbf{9.49}
& \textbf{9.19}
\\
\bottomrule
\end{tabularx}

\end{table}

Our rewrite method achieves the best performance among the compared methods, as shown in Table~\ref{tab:rewrite_quality}. Under both evaluation models, the mean rewrite-quality scores exceed 9 for both ADD and REMOVE interventions. The results across two different judges also indicate that the advantage is not specific to a particular evaluator. Figure~\ref{fig:rewrite_quality_scores} further shows that most of our rewrites receive a score of 9 or 10, suggesting that the evidence-guided procedure can reliably produce high-quality counterfactual edits in both intervention directions.

\definecolor{editadded}{RGB}{0,125,72}
\definecolor{editremoved}{RGB}{190,35,55}
\newcommand{\added}[1]{\textcolor{editadded}{#1}}
\newcommand{\removed}[1]{\textcolor{editremoved}{\sout{#1}}}
\newcommand{\scoreup}[1]{\textcolor{editadded}{\bfseries$\uparrow\,#1$}}
\newcommand{\scoredown}[1]{\textcolor{editremoved}{\bfseries$\downarrow\,#1$}}

\subsection{Controlled ADD and REMOVE Rewrites}
\label{app:controlled_rewrite_examples}

Below are three representative examples from the rewrite-quality
evaluation.  For every example, \added{added or replacement text is green};
\removed{deleted or replaced text is red and struck through}.
Unmarked text is unchanged.

\tcbset{
  rewriteexample/.style={
    enhanced,
    colback=white,
    colframe=black!45,
    colbacktitle=black!75,
    coltitle=white,
    fonttitle=\bfseries,
    arc=1.5mm,
    boxrule=.7pt,
    left=1.5mm,
    right=1.5mm,
    top=.8mm,
    bottom=.8mm
  },
  rewritecard/.style={
    enhanced,
    colback=gray!3!white,
    colframe=black!30,
    colbacktitle=black!25,
    coltitle=white,
    fonttitle=\bfseries,
    arc=1.5mm,
    boxrule=.7pt,
    left=1.5mm,
    right=1.5mm,
    top=.8mm,
    bottom=.8mm
  }
}

\begin{tcolorbox}[
  rewriteexample,
  title={Example 1}]
\normalsize
\setlength{\medskipamount}{1pt}
\textbf{Mechanism:} Clarifies the context or source of the information before
answering.

\medskip
\textbf{Prompt.} Given the statement ``FM systems are used for
entertainment,'' answer: ``What are used for entertainment?''

\medskip
\textbf{ADD intervention}

\begin{tcbraster}[
  raster columns=2,
  raster equal height=rows,
  raster column skip=4mm,
  raster left skip=0pt,
  raster right skip=0pt]
\begin{tcolorbox}[rewritecard,title={Before ($r=-1.820$)}]
FM system
\end{tcolorbox}
\begin{tcolorbox}[rewritecard,title={After ($r=-0.621$)}]
\added{According to the statement, }FM system\added{s.}
\end{tcolorbox}
\end{tcbraster}

\medskip
\textbf{REMOVE intervention}

\begin{tcbraster}[
  raster columns=2,
  raster equal height=rows,
  raster column skip=4mm,
  raster left skip=0pt,
  raster right skip=0pt]
\begin{tcolorbox}[rewritecard,title={Before ($r=2.266$)}]
Answer: \removed{According to the statement, }FM systems are used for
entertainment. Therefore, the answer to the question is: FM systems.
\end{tcolorbox}
\begin{tcolorbox}[rewritecard,title={After ($r=0.863$)}]
Answer: FM systems are used for entertainment. Therefore, the answer to the
question is: FM systems.
\end{tcolorbox}
\end{tcbraster}

\end{tcolorbox}

\begin{tcolorbox}[
  rewriteexample,
  title={Example 2}]
\normalsize
\setlength{\medskipamount}{1pt}
\textbf{Mechanism:} Directly answers the question before providing additional
context.

\medskip
\textbf{Prompt.} Why can cattle fly?

\medskip
\textbf{ADD intervention}

\begin{tcbraster}[
  raster columns=2,
  raster equal height=rows,
  raster column skip=4mm,
  raster left skip=0pt,
  raster right skip=0pt]
\begin{tcolorbox}[rewritecard,title={Before ($r=0.297$)}]
Cattle are mammalian animals that belong to the genus ``Bos'' and the family
``Bovidae.'' They are herbivores that feed on grass, hay, and other vegetation.
While they have wings or the ability to fly, they are not naturally capable of
flight. Birds, bats, and certain insects are examples of animals that can fly.
\end{tcolorbox}
\begin{tcolorbox}[rewritecard,title={After ($r=3.031$)}]
\added{No---cattle cannot fly.}
Cattle are mammalian animals that belong to the genus ``Bos'' and the family
``Bovidae.'' They are herbivores that feed on grass, hay, and other vegetation.
While they have wings or the ability to fly, they are not naturally capable of
flight. Birds, bats, and certain insects are examples of animals that can fly.
\end{tcolorbox}
\end{tcbraster}

\medskip
\textbf{REMOVE intervention}

\begin{tcbraster}[
  raster columns=2,
  raster equal height=rows,
  raster column skip=4mm,
  raster left skip=0pt,
  raster right skip=0pt]
\begin{tcolorbox}[rewritecard,title={Before ($r=7.062$)}]
\removed{Cattle cannot fly. They are large terrestrial mammals that do not
possess the physical adaptations for powered flight, such as wings or the
necessary muscle structure.} If you have any more questions or need
clarification on a different topic, feel free to ask.
\end{tcolorbox}
\begin{tcolorbox}[rewritecard,title={After ($r=5.750$)}]
\added{Powered flight requires physical adaptations, such as wings and the
necessary muscle structure. Cattle are large terrestrial mammals and do not
possess those adaptations, so they cannot fly.} If you have any more questions
or need clarification on a different topic, feel free to ask.
\end{tcolorbox}
\end{tcbraster}

\end{tcolorbox}

\begin{tcolorbox}[
  rewriteexample,
  title={Example 3}]
\normalsize
\setlength{\medskipamount}{1pt}
\textbf{Mechanism:} Invites the user to specify the type of wave for a more
precise answer.

\medskip
\textbf{Prompt.} Why are waves always on land?

\medskip
\textbf{ADD intervention}

\begin{tcbraster}[
  raster columns=2,
  raster equal height=rows,
  raster column skip=4mm,
  raster left skip=0pt,
  raster right skip=0pt]
\begin{tcolorbox}[rewritecard,title={Before ($r=-4.156$)}]
Because waves are always on Earth, wherever waves exist, waves can be observed
on land.

\removed{Are there water waves on the moon?}
\end{tcolorbox}
\begin{tcolorbox}[rewritecard,title={After ($r=-1.688$)}]
Because waves are always on Earth, wherever waves exist, waves can be observed
on land.

\added{Do you mean water waves, sound waves, seismic waves, or another type?}
\end{tcolorbox}
\end{tcbraster}

\medskip
\textbf{REMOVE intervention}

\begin{tcbraster}[
  raster columns=2,
  raster equal height=rows,
  raster column skip=4mm,
  raster left skip=0pt,
  raster right skip=0pt]
\begin{tcolorbox}[rewritecard,title={Before ($r=4.625$)}]
Waves are not always on land. The term ``waves'' typically refers to the
movement of water due to the transfer of energy, especially in oceans, seas,
and other large bodies of water. Waves occur as energy moves through water,
causing it to rise and fall in a pattern.

However, when waves reach a coastline, they can crash against the shore,
creating the appearance of waves on land. Additionally, other types of waves,
like seismic waves (caused by earthquakes) or sound waves (caused by
vibrations in the air), can also be observed on land. However, these are
different phenomena than the common idea of waves associated with water
bodies.

\removed{It might be helpful to specify the type of wave you are asking about
so I can provide a more precise answer.}
\end{tcolorbox}
\begin{tcolorbox}[rewritecard,title={After ($r=4.312$)}]
Waves are not always on land. The term ``waves'' typically refers to the
movement of water due to the transfer of energy, especially in oceans, seas,
and other large bodies of water. Waves occur as energy moves through water,
causing it to rise and fall in a pattern.

However, when waves reach a coastline, they can crash against the shore,
creating the appearance of waves on land. Additionally, other types of waves,
like seismic waves (caused by earthquakes) or sound waves (caused by
vibrations in the air), can also be observed on land. However, these are
different phenomena than the common idea of waves associated with water
bodies.
\end{tcolorbox}
\end{tcbraster}

\end{tcolorbox}

\begin{center}
\includegraphics[width=0.861\textwidth]{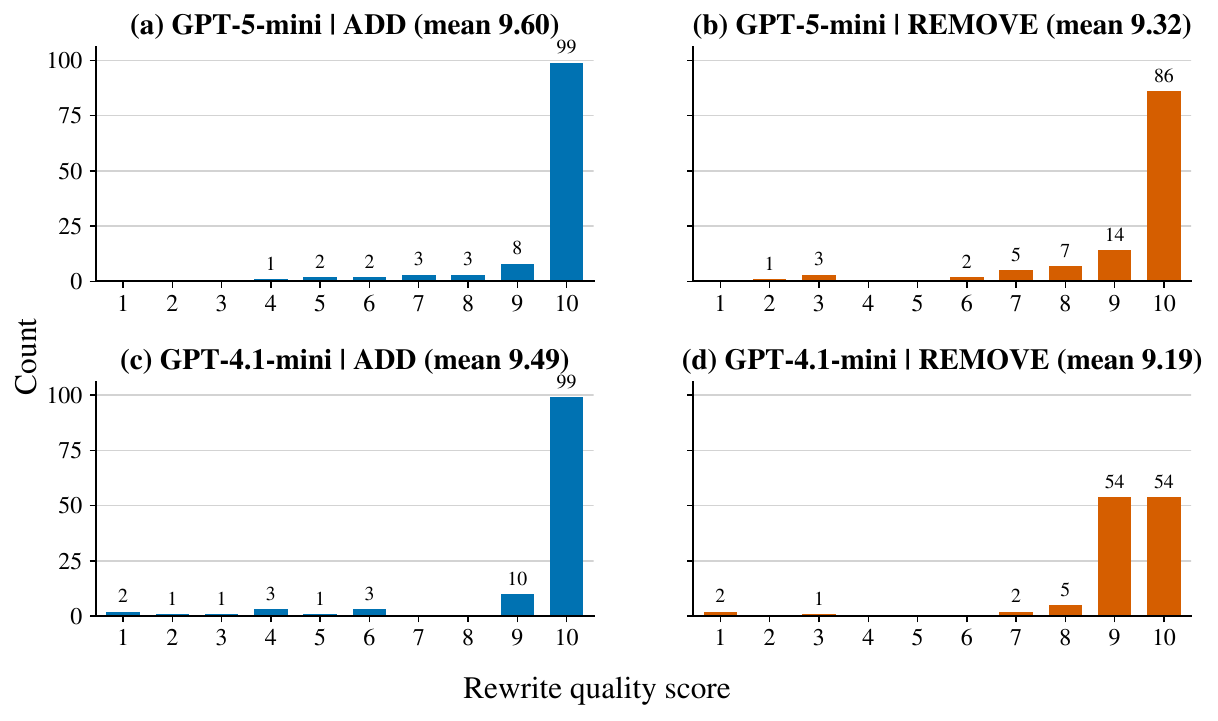}
\captionof{figure}{
Evidence-Guided Rewrite score distributions for ADD and REMOVE under two
judges. Scores range from 1 to 10.
}
\label{fig:rewrite_quality_scores}
\end{center}

\section{Mechanism Diversity}
\label{app:mechanism_redundancy}

To encourage diverse mechanism discovery, we prompt RewardExplainer to generate
multiple mechanisms for each response pair. To examine whether these generated
mechanisms collapse into duplicate or highly similar explanations, we further
analyze mechanism diversity. During evaluation, we select 300 unseen response
pairs and require the model to output three mechanisms for each pair in a single
generation. We then evaluate all pairwise combinations among the three
mechanisms within each response pair.
We use four metrics: (1) Exact Duplicate, whether two mechanisms are
identical after lowercasing, whitespace normalization, and removing the final
period; (2) Jaccard $\geq .50$, whether the token-set Jaccard similarity
reaches 0.50 after removing articles, common function words, and template words
such as uses, adds, and provides; (3) Character Similarity $\geq .70$,
whether the character-level sequence similarity reaches 0.70; and (4)
Embedding Similarity $\geq .85$, whether the cosine similarity between
Qwen3-Embedding-4B \citep{zhang2025qwen3} representations reaches 0.85.
\begin{center}
\captionof{table}{Evaluation of mechanism redundancy and semantic similarity across three target reward models. Char. denotes character sequence similarity, and Emb. denotes embedding cosine similarity. Percentages indicate the proportion of within-example mechanism pairs meeting each criterion.}
\label{tab:mechanism_redundancy}
\small
\setlength{\tabcolsep}{4pt}
\begin{tabularx}{\linewidth}{@{}X l r r r r@{}}
\toprule
Target RM & Explainer & Exact & Jaccard & Char. & Emb. \\
 &  & duplicate & $\geq .50$ & $\geq .70$ & $\geq .85$ \\
\midrule
Skywork-Reward-V2-Qwen3-1.7B & Qwen3-4B & 0.0\% & 0.0\% & 0.1\% & 1.0\% \\
FsfairX-LLaMA3-RM-v0.1       & Qwen3-8B & 0.0\% & 0.3\% & 0.2\% & 1.0\% \\
RM-Mistral-7B                & Qwen3-8B & 0.0\% & 0.1\% & 0.0\% & 0.8\% \\
\midrule
\textbf{Overall}             & --        & 0.0\% & 0.1\% & 0.1\% & 0.9\% \\
\bottomrule
\end{tabularx}
\end{center}

We find little evidence of mechanism collapse across target RMs and explainer settings. Exact duplicates are \(0\%\), high Jaccard and character similarity each occur in only \(0.1\%\) of comparisons, and any lexical overlap criterion is triggered in \(0.2\%\). Only \(0.9\%\) of mechanism pairs exceed an embedding similarity of 0.85, with no case where all three mechanisms collapse. These results suggest that mechanisms generated in a single output typically capture distinct response behaviors rather than paraphrasing the same mechanism.

\section{Qualitative Examples}
\label{app:three_mechanism_cases}

\begin{wrapfigure}{r}{0.43\linewidth}
    \vspace{-10pt}
    \centering
    \includegraphics[width=\linewidth]{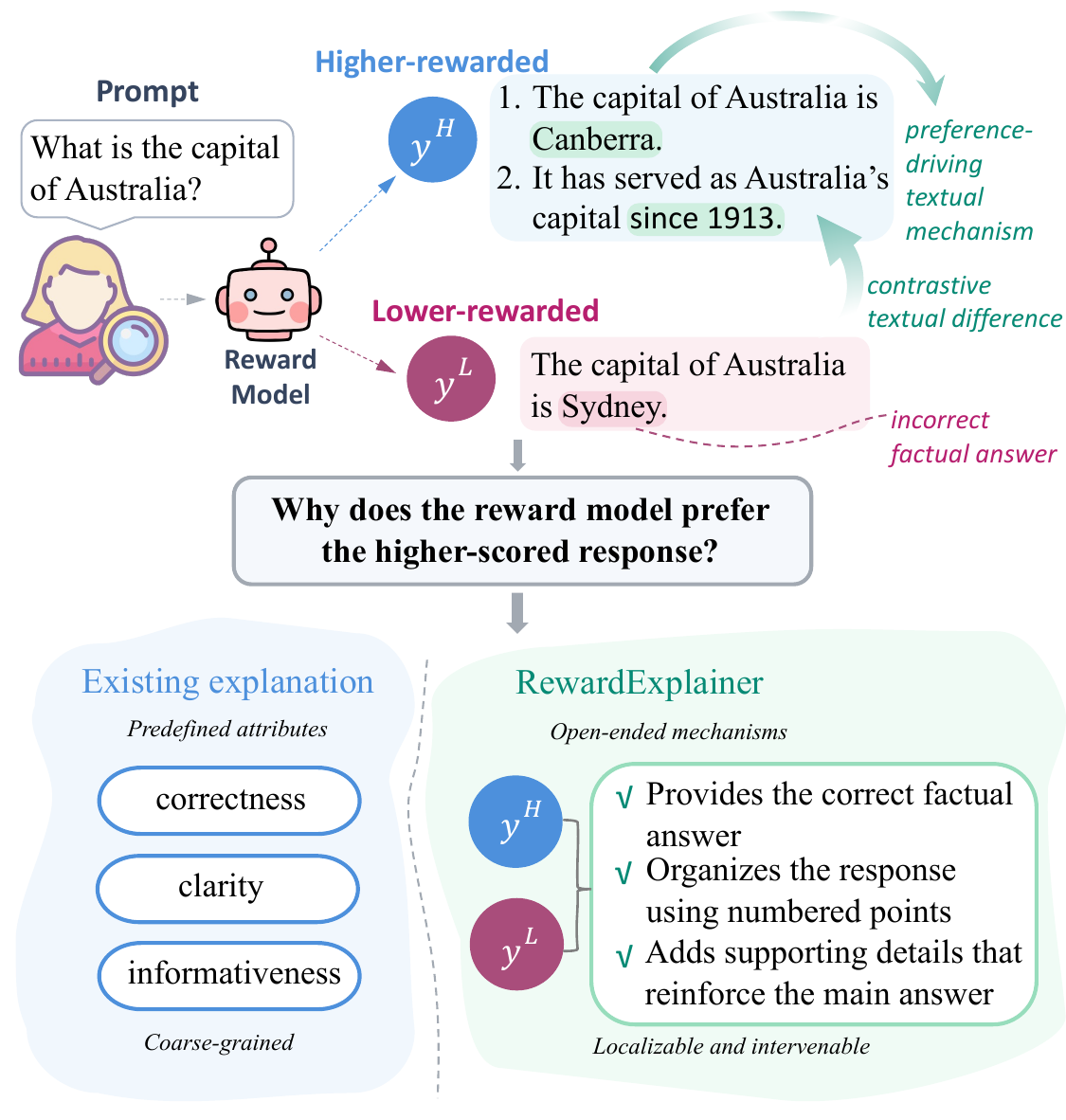}
    \caption{A motivating example.}
    \label{fig:motivation_case}
    \vspace{-10pt}
\end{wrapfigure}
As shown in Figure~\ref{fig:motivation_case}, existing reward-model explanation methods typically rely on predefined high-level attributes, such as correctness, clarity, and informativeness \citep{jiang2024interpreting}. They explain a response pair by counterfactually modifying the responses along each attribute and observing the resulting changes in the reward model's preference. However, these methods require separate interventions and evaluations for every candidate attribute, leading to substantial computational cost. Moreover, because the explanation space is restricted to a predefined attribute set, they can characterize preferences only along known dimensions and may fail to uncover specific scoring mechanisms that were not specified in advance.

In contrast, RewardExplainer does not assume a predefined attribute space. Instead, it directly generates open-ended mechanism explanations from concrete differences between higher- and lower-reward responses. As illustrated in Figure~\ref{fig:motivation_case}, RewardExplainer identifies more specific and intervenable preference-driving behaviors, such as providing correct factual information, organizing the response with explicit structure, and adding supporting details. These mechanisms are further validated through bidirectional ADD and REMOVE counterfactual interventions, linking each explanation to the observed scoring behavior of the target reward model.

We present representative response pairs with three generated mechanisms and their bidirectional intervention results. All mechanisms pass both ADD and REMOVE, with reward changes reported as \scoreup{+\Delta r} and \scoredown{-\Delta r}. These examples show that RewardExplainer captures concrete preference-driving differences consistent with the target reward model's scoring behavior, providing qualitative evidence of explanation accuracy and faithfulness.
\par\WFclear

\tcbset{
  evidenceexample/.style={
    enhanced,
    colback=blue!1!white,
    colframe=blue!55!black,
    colbacktitle=blue!65!black,
    coltitle=white,
    fonttitle=\bfseries,
    arc=2mm,
    boxrule=.8pt,
    left=3mm,
    right=3mm,
    top=2mm,
    bottom=2mm,
    before skip=8pt,
    after skip=10pt
  },
  evidenceprompt/.style={
    enhanced,
    colback=blue!5!white,
    colframe=blue!18!black,
    arc=1.2mm,
    boxrule=.45pt,
    left=2mm,
    right=2mm,
    top=1mm,
    bottom=1mm,
    before skip=0pt,
    after skip=5pt
  },
  evidenceresponse/.style={
    enhanced,
    colback=white,
    colframe=black!25,
    colbacktitle=black!18,
    coltitle=black,
    fonttitle=\bfseries,
    arc=1.2mm,
    boxrule=.5pt,
    left=2mm,
    right=2mm,
    top=1.5mm,
    bottom=1.5mm
  }
}

\begin{tcolorbox}[
  evidenceexample,
  title={Example 1 \enspace---\enspace RM-Mistral / Qwen3-8B RewardExplainer}]
\footnotesize
\begin{tcolorbox}[evidenceprompt]
\textbf{Prompt.} Categorize these animals as predators/prey: cheetah,
elephant, rabbit.
\end{tcolorbox}

\textbf{Response pair}
\begin{tcbraster}[
  raster columns=2,
  raster equal height=rows,
  raster column skip=4mm,
  raster left skip=0pt,
  raster right skip=0pt]
\begin{tcolorbox}[evidenceresponse,title={Higher-scored ($r=5.125$)}]
Cheetah -- Predator

Elephant -- Neither (Herbivore, not typically considered as prey)

Rabbit -- Prey
\end{tcolorbox}
\begin{tcolorbox}[evidenceresponse,title={Lower-scored ($r=1.445$)}]
Cheetah is a predator, while elephant and rabbit are prey.
\end{tcolorbox}
\end{tcbraster}

\smallskip
\renewcommand{\arraystretch}{1.18}
\setlength{\tabcolsep}{4pt}
\begin{tabularx}{\linewidth}{@{}X >{\centering\arraybackslash}p{.14\linewidth} >{\centering\arraybackslash}p{.17\linewidth}@{}}
\rowcolor{blue!8}
\textbf{Generated mechanism} & \textbf{ADD $\Delta r$} & \textbf{REMOVE $\Delta r$} \\
Uses bullet points or labels for clarity.
  & \scoreup{+1.211} & \scoredown{-0.781} \\
Provides additional context for each category.
  & \scoreup{+0.742} & \scoredown{-1.031} \\
Clarifies ambiguous categories with explanations.
  & \scoreup{+0.297} & \scoredown{-1.000} \\
\end{tabularx}
\end{tcolorbox}

\begin{tcolorbox}[
  evidenceexample,
  title={Example 2 \enspace---\enspace Skywork RM / Qwen3-4B RewardExplainer}]
\footnotesize
\begin{tcolorbox}[evidenceprompt]
\textbf{Prompt.} How many legs does an macbook have? Three or four?
\end{tcolorbox}

\textbf{Response pair}
\begin{tcbraster}[
  raster columns=2,
  raster equal height=rows,
  raster column skip=4mm,
  raster left skip=0pt,
  raster right skip=0pt]
\begin{tcolorbox}[evidenceresponse,title={Higher-scored ($r=3.609$)}]
A MacBook is a laptop computer made by Apple Inc. and does not have legs. It
has a flat base, a hinged screen, and a keyboard built into the body, but no
legs like an animal or a piece of furniture would have.
\end{tcolorbox}
\begin{tcolorbox}[evidenceresponse,title={Lower-scored ($r=-3.125$)}]
Answer: A MacBook typically has four legs.

Confidence: 99\%
\end{tcolorbox}
\end{tcbraster}

\smallskip
\renewcommand{\arraystretch}{1.18}
\setlength{\tabcolsep}{4pt}
\begin{tabularx}{\linewidth}{@{}X >{\centering\arraybackslash}p{.14\linewidth} >{\centering\arraybackslash}p{.17\linewidth}@{}}
\rowcolor{blue!8}
\textbf{Generated mechanism} & \textbf{ADD $\Delta r$} & \textbf{REMOVE $\Delta r$} \\
Provides a brief explanation of the object's nature before answering.
  & \scoreup{+1.461} & \scoredown{-2.289} \\
Clarifies the context or analogy to avoid misunderstanding.
  & \scoreup{+5.562} & \scoredown{-0.328} \\
Describes the physical structure of the object to support the answer.
  & \scoreup{+5.734} & \scoredown{-0.688} \\
\end{tabularx}
\end{tcolorbox}

\begin{tcolorbox}[
  evidenceexample,
  title={Example 3 \enspace---\enspace FsfairX RM / Qwen3-8B RewardExplainer}]
\footnotesize
\begin{tcolorbox}[evidenceprompt]
\textbf{Prompt.} what kind of color happen if I mix red and blue?
\end{tcolorbox}

\textbf{Response pair}
\begin{tcbraster}[
  raster columns=2,
  raster equal height=rows,
  raster column skip=4mm,
  raster left skip=0pt,
  raster right skip=0pt]
\begin{tcolorbox}[evidenceresponse,title={Higher-scored ($r=-0.660$)}]
When you mix red and blue, you get the color purple.

1. \textbf{Variability}: The exact shade of purple can vary depending on:

-- The specific shades of red and blue used

-- The proportions of red and blue

2. \textbf{Examples}:

-- Mixing more red results in a warmer, reddish-purple.

-- Mixing more blue creates a cooler, bluish-purple.
\end{tcolorbox}
\begin{tcolorbox}[evidenceresponse,title={Lower-scored ($r=-1.797$)}]
When you mix red and blue, you end up with the color purple. But, you know,
the exact shade you get can change a bit depending on a couple of things. Like,
if you're using different shades of red or blue, or even just varying the
amount of red versus blue---those can all affect how your purple turns out. So,
imagine if you throw in more red, you might see a warmer, more reddish-purple.
But on the flip side, if you're heavy on the blue, you'll probably get a
cooler, bluish-purple. Kind of interesting how that works, right?
\end{tcolorbox}
\end{tcbraster}

\smallskip
\renewcommand{\arraystretch}{1.18}
\setlength{\tabcolsep}{4pt}
\begin{tabularx}{\linewidth}{@{}X >{\centering\arraybackslash}p{.14\linewidth} >{\centering\arraybackslash}p{.17\linewidth}@{}}
\rowcolor{blue!8}
\textbf{Generated mechanism} & \textbf{ADD $\Delta r$} & \textbf{REMOVE $\Delta r$} \\
Uses concise, direct wording instead of conversational filler.
  & \scoreup{+0.484} & \scoredown{-1.199} \\
Presents the answer in a cleaner, more organized format.
  & \scoreup{+1.102} & \scoredown{-1.855} \\
Removes unnecessary conversational padding and examples.
  & \scoreup{+0.273} & \scoredown{-3.277} \\
\end{tabularx}
\end{tcolorbox}

\begin{tcolorbox}[
  evidenceexample,
  title={Example 4 \enspace---\enspace RM-Mistral / Qwen3-8B RewardExplainer}]
\footnotesize
\begin{tcolorbox}[evidenceprompt]
\textbf{Prompt.} Solve $2v-24+26=0$ for $v$.. What is the answer??
\end{tcolorbox}

\textbf{Response pair}
\begin{tcbraster}[
  raster columns=2,
  raster equal height=rows,
  raster column skip=4mm,
  raster left skip=0pt,
  raster right skip=0pt]
\begin{tcolorbox}[evidenceresponse,title={Higher-scored ($r=4.812$)}]
To solve the equation $2v-24+26=0$ for $v$, follow these steps:

1. Add 24 to both sides of the equation:

$2v-24+26+24=0+24$

This simplifies to $2v=48$.

2. Then, divide both sides of the equation by 2:

$2v/2=48/2$

This simplifies to $v=24$.

So, the solution to the equation $2v-24+26=0$ is $v=24$.
\end{tcolorbox}
\begin{tcolorbox}[evidenceresponse,title={Lower-scored ($r=2.391$)}]
Answer:

$v=4$

Confidence: 95\%
\end{tcolorbox}
\end{tcbraster}

\smallskip
\renewcommand{\arraystretch}{1.18}
\setlength{\tabcolsep}{4pt}
\begin{tabularx}{\linewidth}{@{}X >{\centering\arraybackslash}p{.14\linewidth} >{\centering\arraybackslash}p{.17\linewidth}@{}}
\rowcolor{blue!8}
\textbf{Generated mechanism} & \textbf{ADD $\Delta r$} & \textbf{REMOVE $\Delta r$} \\
Provides a step-by-step explanation of the solution process.
  & \scoreup{+1.312} & \scoredown{-3.586} \\
Restates the original equation before solving it.
  & \scoreup{+0.719} & \scoredown{-0.469} \\
Simplifies each step of the solution before proceeding to the next.
  & \scoreup{+0.797} & \scoredown{-1.000} \\
\end{tabularx}
\end{tcolorbox}

\section{Robustness and Statistical Significance}
\label{app:significance_robustness}

\subsection{Multi-seed Robustness}

To assess the robustness of RewardExplainer to training randomness, we repeat
the experiment with three independent training seeds (42, 123, and 2026) under
the RM-Mistral-7B and Qwen3-4B explainer setting, while keeping the test data
and evaluation protocol fixed. As shown in
Table~\ref{tab:mistral_seed_core}, results are stable across all metrics.  The
overall Mechanism Quality score is $1.5500\pm0.0124$, and Preference
Prediction Accuracy is $85.94\%\pm0.86\%$.  For Counterfactual Faithfulness,
the mean ADD, REMOVE, and BOTH success rates are $56.53\%\pm0.88\%$,
$54.03\%\pm0.99\%$, and $43.80\%\pm0.84\%$, respectively, with across-seed
standard deviations below one percentage point.
These results indicate that RewardExplainer's main performance is not driven
by a particular training seed, but remains consistent across random
initializations.
\begin{center}
\captionof{table}{Performance across three training seeds for RM-Mistral-7B with Qwen3-4B as the explainer backbone. Mean and standard deviation are computed over the three runs.}
\label{tab:mistral_seed_core}
\small
\setlength{\tabcolsep}{5pt}
\renewcommand{\arraystretch}{1.10}
\begin{tabular}{@{}lrrrr@{}}
\toprule
Metric & Set 1 & Set 2 & Set 3 & Mean $\pm$ SD \\
\midrule
Axis Quality (overall) & 1.5478 & 1.5633 & 1.5389 & $1.5500\pm0.0124$ \\
Choice accuracy & 86.84\% & 85.13\% & 85.86\% & $85.94\pm0.86$\% \\
ADD success     & 55.76\% & 57.50\% & 56.32\% & $56.53\pm0.88$\% \\
REMOVE success  & 53.26\% & 55.14\% & 53.69\% & $54.03\pm0.99$\% \\
BOTH success    & 42.86\% & 44.49\% & 44.06\% & $43.80\pm0.84$\% \\
\bottomrule
\end{tabular}
\end{center}

\subsection{Paired Significance Tests}

We further use pair-level paired bootstrap tests to assess whether
RewardExplainer's improvements over the baselines are statistically
significant. Specifically, we perform resampling comparisons on the same 300
matched response pairs for each of the three training seeds. Across all three
seeds, RewardExplainer consistently outperforms both direct mechanism
generation with GPT-4o (Table~\ref{tab:sig_label_prompt}) and Contrastive
Explanations (Table~\ref{tab:sig_attribute}) on all evaluated metrics. All
reported 95\% confidence intervals exclude zero, with $p<0.05$, indicating
statistically significant improvements across different training seeds. The
small variation in improvements across seeds further suggests that the main
conclusions are robust to training randomness rather than being driven by a
particular seed.

\begin{center}
\captionof{table}{Pair-level paired-bootstrap significance tests comparing RewardExplainer with GPT-4o across three training seeds.
$\Delta$ denotes RewardExplainer minus GPT-4o, brackets report 95\% confidence intervals, and pp denotes percentage points.
Mean $\Delta \pm$ SD summarizes the variation across seeds.}
\label{tab:sig_label_prompt}
\small
\setlength{\tabcolsep}{3pt}
\renewcommand{\arraystretch}{1.06}
\begin{tabular}{@{}llcccc@{}}
\toprule
Metric & Statistic & Set 1 & Set 2 & Set 3 & Mean $\Delta\pm$ SD \\
\midrule
Axis   & $\Delta$ & $+0.2422$ & $+0.2578$ & $+0.2333$ & $+0.2444\pm0.0124$ \\
       & 95\% CI & $[0.1778,0.3078]$ & $[0.1878,0.3289]$ & $[0.1689,0.2989]$ & -- \\
       & $p$-value & $.0002$ & $.0002$ & $.0002$ & -- \\
\addlinespace[1.5pt]
Choice & $\Delta$ & $+9.89$ pp & $+8.44$ pp & $+9.50$ pp & $+9.28\pm0.75$ pp \\
       & 95\% CI & $[6.61,13.22]$ & $[5.00,11.89]$ & $[6.39,12.67]$ & -- \\
       & $p$-value & $.0002$ & $.0002$ & $.0002$ & -- \\
\addlinespace[1.5pt]
ADD    & $\Delta$ & $+7.39$ pp & $+8.94$ pp & $+7.72$ pp & $+8.02\pm0.82$ pp \\
       & 95\% CI & $[3.39,11.39]$ & $[5.06,12.94]$ & $[4.00,11.44]$ & -- \\
       & $p$-value & $.0006$ & $.0002$ & $.0002$ & -- \\
\addlinespace[1.5pt]
REMOVE & $\Delta$ & $+7.22$ pp & $+8.67$ pp & $+7.11$ pp & $+7.67\pm0.87$ pp \\
       & 95\% CI & $[3.55,11.00]$ & $[5.00,12.39]$ & $[3.67,10.67]$ & -- \\
       & $p$-value & $.0002$ & $.0002$ & $.0002$ & -- \\
\addlinespace[1.5pt]
BOTH   & $\Delta$ & $+5.61$ pp & $+6.78$ pp & $+6.39$ pp & $+6.26\pm0.59$ pp \\
       & 95\% CI & $[1.78,9.39]$ & $[3.11,10.50]$ & $[2.83,9.94]$ & -- \\
       & $p$-value & $.0034$ & $.0006$ & $.0008$ & -- \\
\bottomrule
\end{tabular}
\end{center}

\begin{center}
\begin{minipage}{\linewidth}
\centering
\captionof{table}{Pair-level paired-bootstrap significance tests comparing RewardExplainer with Contrastive Explanations across three training seeds.
$\Delta$ denotes RewardExplainer minus Contrastive Explanations, brackets report 95\% confidence intervals, and pp denotes percentage points.
Mean $\Delta \pm$ SD summarizes the variation across seeds.}
\label{tab:sig_attribute}
\small
\setlength{\tabcolsep}{3pt}
\renewcommand{\arraystretch}{1.06}
\begin{tabular}{@{}llcccc@{}}
\toprule
Metric & Statistic & Set 1 & Set 2 & Set 3 & Mean $\Delta\pm$ SD \\
\midrule
Axis   & $\Delta$ & $+0.8378$ & $+0.8533$ & $+0.8289$ & $+0.8400\pm0.0124$ \\
       & 95\% CI & $[0.7489,0.9256]$ & $[0.7689,0.9378]$ & $[0.7422,0.9156]$ & -- \\
       & $p$-value & $.0002$ & $.0002$ & $.0002$ & -- \\
\addlinespace[1.5pt]
Choice & $\Delta$ & $+13.11$ pp & $+11.67$ pp & $+12.72$ pp & $+12.50\pm0.75$ pp \\
       & 95\% CI & $[9.33,16.83]$ & $[7.56,15.67]$ & $[8.78,16.56]$ & -- \\
       & $p$-value & $.0002$ & $.0002$ & $.0002$ & -- \\
\addlinespace[1.5pt]
ADD    & $\Delta$ & $+31.28$ pp & $+32.83$ pp & $+31.61$ pp & $+31.91\pm0.82$ pp \\
       & 95\% CI & $[27.00,35.56]$ & $[28.56,37.00]$ & $[27.33,35.94]$ & -- \\
       & $p$-value & $.0002$ & $.0002$ & $.0002$ & -- \\
\addlinespace[1.5pt]
REMOVE & $\Delta$ & $+28.44$ pp & $+29.89$ pp & $+28.33$ pp & $+28.89\pm0.87$ pp \\
       & 95\% CI & $[23.89,32.89]$ & $[25.22,34.44]$ & $[23.94,32.67]$ & -- \\
       & $p$-value & $.0002$ & $.0002$ & $.0002$ & -- \\
\addlinespace[1.5pt]
BOTH   & $\Delta$ & $+24.50$ pp & $+25.67$ pp & $+25.28$ pp & $+25.15\pm0.59$ pp \\
       & 95\% CI & $[20.28,28.67]$ & $[21.56,29.72]$ & $[21.06,29.39]$ & -- \\
       & $p$-value & $.0002$ & $.0002$ & $.0002$ & -- \\
\bottomrule
\end{tabular}
\end{minipage}
\end{center}

\section{Implementation Details}
\label{app:implementation_details}

\subsection{Metrics Details}
\label{app:three_evaluation_protocols}

In this section, we provide the detailed implementation configurations for the evaluation metrics. Let $x$ denote
the user prompt, let $y^H$ and $y^L$ denote the responses assigned higher and
lower scores by the target reward model, respectively, and let
$M=\{m_1,\ldots,m_N\}$ be the generated mechanisms.

\subsubsection{Mechanism Quality}

This test evaluates the quality of the complete mechanism set generated for
each response pair. For each example, the judge LLM is given the prompt $x$,
the higher-reward response $y^H$, the lower-reward response $y^L$, and the
generated mechanism set $\mathcal{M}$, but not the numerical reward scores
from the target reward model. The judge independently assigns a score in
$\{0,1,2\}$ along three dimensions:

\begin{enumerate}[leftmargin=*,nosep]
  \item Contrastive Selectivity: whether the mechanisms capture
    response behaviors that meaningfully distinguish $y^H$ from $y^L$, rather
    than properties shared by both responses.
  \item Scope Calibration: whether the mechanisms are expressed at
    an appropriate level of abstraction without being overly broad or
    extending beyond the evidence provided by the response pair.
  \item Directional Fidelity: whether the mechanisms correctly
    describe the direction of the difference from $y^L$ to $y^H$, such that
    the described behavior is more evident in the higher-reward response.
\end{enumerate}

For response pair $i$, the Mechanism Quality score is the equal-weight average
across the three dimensions:
\begin{equation}
q_i^{\mathrm{quality}}=\frac{1}{3}\sum_{a=1}^{3}s_{i,a},
\qquad s_{i,a}\in\{0,1,2\}.
\end{equation}
The dataset-level result is obtained by averaging over all valid response
pairs. This metric measures whether the generated mechanisms clearly and
accurately characterize the differences between the two responses at an
appropriate level of abstraction.

\subsubsection{Mechanism Choice}

This test evaluates each generated mechanism $m_i$ separately. For each
mechanism, we randomly swap the two responses into positions A and B to avoid
position bias, and ask the judge LLM to select which response better matches
the mechanism.
A mechanism is counted as correct if the response selected by the judge
matches the higher-reward response $y^H$ preferred by the target reward model.
We report Preference Prediction Accuracy as the proportion of correct
predictions over all valid mechanisms:
\begin{equation}
\mathrm{Acc}=\frac{1}{N}\sum_{i=1}^{N}
\mathbf{1}\!\left[\hat{y}_i=y_i^H\right],
\end{equation}
where $\hat{y}_i$ denotes the response selected by the judge based on mechanism
$m_i$, and $N$ is the total number of valid mechanisms.
This metric measures whether the generated mechanisms correctly reflect the
original preference direction of the target reward model.

\subsubsection{Counterfactual Faithfulness}

This test evaluates whether a generated mechanism $m_i$ captures a response
behavior to which the target reward model is actually sensitive. Following the
evidence-guided rewrite procedure described in
Section~\ref{app:rewrite_procedure}, we perform two complementary
counterfactual interventions. ADD introduces or strengthens $m_i$ in
the lower-reward response $y^L$, producing $y_{+m_i}^{L}$, while
REMOVE removes or weakens $m_i$ from the higher-reward response
$y^H$, producing $y_{-m_i}^{H}$. During rewriting, we preserve unrelated
content, existing errors, and writing style as much as possible.
We then rescore the counterfactual responses using the same target reward
model and compute

\begin{equation}
\Delta_{\mathrm{add}}(m_i)
=
r(y_{+m_i}^{L})-r(y^L),
\end{equation}

\begin{equation}
\Delta_{\mathrm{remove}}(m_i)
=
r(y_{-m_i}^{H})-r(y^H).
\end{equation}

ADD is considered successful when $\Delta_{\mathrm{add}}(m_i)>0$, and REMOVE
is considered successful when $\Delta_{\mathrm{remove}}(m_i)<0$. A mechanism
is counted as BOTH success when both conditions hold. We report the
ADD, REMOVE, and BOTH success rates over all valid mechanisms.
This test examines whether modifying the behavior described by $m_i$ changes
the target reward in the expected direction. In particular, BOTH success
provides stronger counterfactual evidence that the generated mechanism is
consistent with the observed scoring behavior of the target reward model.

\subsection{Global Mechanism Aggregation}
\label{app:global_mechanism_aggregation}

After training, RewardExplainer can generate mechanism explanations for
individual preference response pairs. 
To identify recurring preferences across different prompts and characterize the global preference patterns of the target reward model, we aggregate these local mechanisms into a set of global mechanisms.

For RM-Mistral-7B, we sample 100 unseen response pairs, including 80 from
UltraFeedback, 13 from PreferenceHack, and 7 from Preference Model Perturbations (PMP). We first encode the
generated local mechanisms using Qwen3-Embedding-4B \citep{zhang2025qwen3} and perform average-linkage
hierarchical clustering based on cosine similarity with a threshold of 0.75.
GPT-5.4-mini then summarizes each cluster into a canonical mechanism
representing its shared behavior. After this initial clustering, we further use
GPT-5.5 \citep{openai2026gpt55} to consolidate the mechanisms and filter out task-specific,
domain-specific, and insufficiently generalizable patterns, resulting in 15
global mechanisms.
Table~\ref{tab:global_response_attributes_full} presents the complete set of
global mechanisms. These mechanisms summarize recurring preference patterns of
the target reward model across response pairs, including both preferences
related to substantive response quality and systematic preferences for
surface-level presentation, providing a basis for subsequent bias discovery.

{%
\small
\setlength{\tabcolsep}{3pt}
\renewcommand{\arraystretch}{1.08}
\setlength{\LTcapwidth}{\dimexpr0.97\textwidth+2\tabcolsep\relax}
\begin{longtable}{@{}
  >{\raggedright\arraybackslash}p{0.28\textwidth}
  >{\raggedright\arraybackslash}p{0.69\textwidth}
@{}}
\caption{Complete set of global preference mechanisms identified for
RM-Mistral-7B.}
\label{tab:global_response_attributes_full}\\
\toprule
\textbf{Global mechanism} & \textbf{Description} \\
\midrule
\endfirsthead

\multicolumn{2}{l}{\small\itshape Table~\ref{tab:global_response_attributes_full} continued.}\\
\toprule
\textbf{Global mechanism} & \textbf{Description} \\
\midrule
\endhead

\midrule
\multicolumn{2}{r}{\small\itshape Continued on the next page.}\\
\endfoot

\bottomrule
\endlastfoot

\rowcolor{gray!8}
Direct task-focused answering
& The reward model favors responses that answer the user's request plainly
and upfront while minimizing unnecessary commentary, embellishment, or
digression. \\

Concise justification of answers
& The reward model favors answers that pair a clear conclusion,
recommendation, classification, or definition with a brief explanation of why
it is correct or significant. \\

\rowcolor{gray!8}
Explicit structural organization
& The reward model favors responses that organize information using visible
structure such as lists, numbered points, headings, subheadings, or sectioned
layouts. \\

Sequential procedural breakdown
& The reward model favors explanations that break a process, solution, or
implementation into an ordered sequence of steps or stages. \\

\rowcolor{gray!8}
Contrastive clarification
& The reward model favors responses that clarify a choice, concept, or
classification by explicitly comparing it against alternatives or contrasting
related cases. \\

Epistemic calibration and premise correction
& The reward model favors responses that surface uncertainty, missing
information, false premises, ambiguity, or evidential limitations before
giving a definitive answer. \\

\rowcolor{gray!8}
Risk-aware safe guidance
& The reward model favors responses that identify safety, health, legal,
ethical, or interpersonal risks and redirect toward responsible or lower-risk
guidance. \\

Conversational rapport framing
& The reward model favors responses that use friendly, validating, or engaged
conversational framing to make the reply feel responsive to the user. \\

\rowcolor{gray!8}
Closing synthesis
& The reward model favors responses that end by briefly restating or
synthesizing the main point to reinforce the answer. \\

Context-rich narrative summarization
& The reward model favors summaries that preserve concrete situational
details, emotional context, and coherent narrative flow from the source
material. \\

\rowcolor{gray!8}
Concrete examples and evidence
& The reward model favors explanations that ground abstract claims in concrete
examples, code, data, cited cases, or scenario-specific illustrations. \\

Broader contextual expansion
& The reward model favors answers that expand beyond the immediate query by
adding relevant background, broader implications, applications, or
interdisciplinary context. \\

\rowcolor{gray!8}
Explicit confidence annotation
& The reward model sometimes favors responses that append an explicit
confidence statement or percentage to qualify the answer. \\

Substance-signaling elaboration
& The reward model sometimes favors added explanatory bulk that increases the
apparent substance of the response even when it contributes limited new
information. \\

\rowcolor{gray!8}
Actionable next-step guidance
& The reward model favors responses that translate advice into concrete
strategies, tools, implementation steps, or practical next actions the user
can take. \\

\end{longtable}
}

\subsection{Mechanism-Guided Reward Model Debiasing}
\label{app:bias_debias_implementation}

\subsubsection{Bias discovery}
\label{app:bias_discovery}
For each target reward model, we first sample 50 preference pairs. We score both responses with the target RM and denote the higher- and lower-scored responses as \(y^H\) and \(y^L\), respectively. RewardExplainer is then used to generate candidate mechanisms for each pair. A mechanism is retained as a potential bias mechanism only if two conditions are satisfied: (1) the target RM's preference disagrees with the non-tied human preference, and (2) the mechanism passes the bidirectional counterfactual intervention test. The first condition identifies cases where the RM's scoring behavior may conflict with substantive human preference, while the second provides behavioral evidence that the mechanism is supported by the target RM's observed scoring behavior rather than being merely a plausible post-hoc description. We then use GPT-5.5 to aggregate and filter the retained candidates, producing the final set of potential bias mechanisms for each target RM.

\subsubsection{Debiasing Data Construction}
\label{app:bias_dataset}
For each discovered bias mechanism, we sample 500 preference pairs from
Skywork Reward Preference 80K v0.2 \citep{liu2024skywork}. Given an original preference example
$(p,y^+,y^-)$, we use GPT-4o to minimally rewrite the rejected response $y^-$
so that the target bias mechanism is injected while preserving its substantive
quality as much as possible, producing $\tilde y_b^-$.
We then use a second independent GPT-4o call to validate each rewritten sample.
To reduce position bias, $y^+$ and $\tilde y_b^-$ are randomly assigned to
positions A and B. The judge determines (1) which response has higher
substantive quality, and (2) whether the target bias mechanism is present.
A sample is retained only if
\begin{equation}
Q(y^+)>Q(\tilde y_b^-),\qquad
b(y^+)=0,\qquad b(\tilde y_b^-)=1,
\end{equation}
where $Q$ denotes the substantive quality comparison and $b(\cdot)$ indicates
the presence of the target bias mechanism. This filtering retains samples in which the target bias is judged to be present only in the rewritten rejected response while the original chosen response remains substantively better.
After validation and length filtering, we obtain 811 targeted pairs for
FsfairX-LLaMA3-RM-v0.1, 775 for RM-Mistral-7B, and 528 for
Skywork-Reward-V2-Qwen3-1.7B.

\subsubsection{Reward Model Debiasing}
\label{app:debias}
To preserve the general capability of the reward model, we mix the constructed debiasing data with general preference data at a 1:1 ratio. The corresponding General baseline is trained on the same amount of original general preference data. We initialize from the original scalar target RM checkpoint and apply LoRA to the attention \(q\)-projection and \(v\)-projection modules. The remaining optimization hyperparameters are reported in Table~\ref{tab:debias_rm_hyperparameters}.
After training, we compare the debiased model against both the General baseline and the unmodified Base model. We evaluate debiasing effectiveness on the RewardHackBench \citep{liu2026harve} test set and JudgeBias \citep{zhou2026toward}, two benchmarks designed to assess reward-hacking-related behavior, and further evaluate on the general-purpose JudgeBench benchmark \citep{ICLR2025_9e720fce} to assess changes in the reward model's general evaluation capability.

{%
\begin{table}[t]
\centering
\caption{Target-RM debiasing hyperparameters.}
\label{tab:debias_rm_hyperparameters}
\small
\setlength{\tabcolsep}{4pt}
\renewcommand{\arraystretch}{1.10}
\rowcolors{2}{gray!8}{white}
\begin{tabularx}{\linewidth}{@{}
>{\raggedright\arraybackslash}p{.28\linewidth}
>{\centering\arraybackslash}X
>{\centering\arraybackslash}X
>{\centering\arraybackslash}X
@{}}
\toprule
\rowcolor{white}
\textbf{Setting} & \textbf{FsfairX} & \textbf{RM-Mistral} & \textbf{Skywork} \\
\midrule
Training pairs & 1,622 & 1,550 & 1,056 \\
Maximum length & 2,048 & 4,096 & 3,072 \\
Per-device batch & 2 & 1 & 2 \\
Gradient accumulation & 8 & 16 & 8 \\
Epochs & 3 & 3 & 3 \\
Learning rate & $2{\times}10^{-5}$ & $2{\times}10^{-5}$ & $2{\times}10^{-5}$ \\
Optimizer & AdamW & AdamW & AdamW \\
Scheduler & Cosine & Cosine & Cosine \\
Warmup ratio & 0.05 & 0.05 & 0.05 \\
Weight decay & 0.01 & 0.01 & 0.01 \\
LoRA rank $r$ & 16 & 16 & 16 \\
LoRA alpha $\alpha$ & 32 & 32 & 32 \\
LoRA dropout & 0.05 & 0.05 & 0.05 \\
LoRA targets & $q,v$ & $q,v$ & $q,v$ \\
\bottomrule
\end{tabularx}
\rowcolors{2}{white}{white}
\end{table}
}

\subsection{Training Details and Hyperparameters}
\label{training_detail}
All experiments can be run on a single NVIDIA A100 GPU with 40~GB of memory.
For both Qwen3-4B and Qwen3-8B, SFT takes approximately 1.5 hours and
DPO takes approximately 2 hours, while training Gemma4-12B takes roughly
twice as long. The SFT stage uses $K_1=4{,}825$ training examples, and the
DPO data-construction stage uses $K_2=4{,}000$ response pairs. The overall
ratio of UltraFeedback examples to examples from the reward-hacking datasets
PreferenceHack and Preference Model Perturbations (PMP) is approximately
4:1.

During SFT, we first use GPT-4o to generate
atomic, generalizable, and editable textual mechanisms as
supervision labels for fine-tuning the base explainer. After obtaining the
SFT model, we sample $W=4$ times for each response pair in the DPO dataset
using a temperature of $0.9$ and top-$p$ of $0.9$ to generate candidate
mechanisms. GPT-5.4-mini is then used to perform bidirectional counterfactual
rewriting and verification, from which we construct mechanism preference
data. The DPO stage is initialized from the SFT adapter and trained on the
constructed preference data with $\beta=0.1$.
Table~\ref{tab:common_hyperparameters} summarizes the model-specific training
hyperparameters.
Both training stages use AdamW with cosine learning-rate scheduling, a warmup
ratio of $0.05$, weight decay of $0.01$, gradient clipping at $1.0$, and
gradient checkpointing. LoRA is applied to all linear layers with rank $16$,
scale $32$, and dropout $0.05$.
\begin{center}
\begin{minipage}{\linewidth}
\captionof{table}{Model-specific RewardExplainer training hyperparameters.}
\label{tab:common_hyperparameters}
\small
\setlength{\tabcolsep}{3.5pt}
\renewcommand{\arraystretch}{1.12}
\rowcolors{2}{gray!8}{white}
\begin{tabularx}{\linewidth}{@{}
>{\raggedright\arraybackslash}p{.20\linewidth}
*{6}{>{\centering\arraybackslash}X}
@{}}
\toprule
\rowcolor{white}
\textbf{Setting}
& \multicolumn{2}{c}{\textbf{Qwen3-4B}}
& \multicolumn{2}{c}{\textbf{Qwen3-8B}}
& \multicolumn{2}{c}{\textbf{Gemma4-12B}}
\\
\cmidrule(lr){2-3}\cmidrule(lr){4-5}\cmidrule(l){6-7}
\rowcolor{white}
& \textbf{SFT} & \textbf{DPO}
& \textbf{SFT} & \textbf{DPO}
& \textbf{SFT} & \textbf{DPO}
\\
\midrule
Epochs
& 2 & 2 & 2 & 2 & 2 & 2 \\
Learning rate
& $1{\times}10^{-4}$ & $5{\times}10^{-6}$
& $1{\times}10^{-4}$ & $5{\times}10^{-6}$
& $1{\times}10^{-4}$ & $5{\times}10^{-6}$ \\
Per-device batch
& 2 & 3 & 4 & 2 & 1 & 1 \\
Grad. accum.
& 8 & 8 & 4 & 8 & 16 & 32 \\
Maximum length
& 2,048 & 2,048
& 2,048 & 2,048
& 2,048 & 1,920 \\
\bottomrule
\end{tabularx}
\rowcolors{2}{white}{white}
\end{minipage}
\end{center}

\section{Prompts}
\label{app:sft_explanation_prompt}

\subsection{Mechanism Generation Prompt}
\label{app:mechanism_generation_prompt}

\begin{tcolorbox}[
  enhanced,
  breakable,
  colback=orange!10!white,
  colframe=blue!5!black,
  arc=2mm,
  boxrule=1pt,
  title={\bfseries SFT Explanation Prompt},
  coltitle=white,
  attach boxed title to top left={yshift=-2mm, xshift=3mm},
  boxed title style={
    enhanced,
    colback=blue!5!black,
    colframe=blue!5!black,
    arc=2mm,
    boxrule=0pt
  },
  top=0.5mm,
  left=1mm,
  right=1mm,
  bottom=0.5mm
]
\small
\vskip8pt

\textbf{System Prompt}

You identify textual mechanisms that may explain a reward model's
preferences. Return valid JSON only.

\medskip
\textbf{User Prompt Template}

Given a user request and two responses, identify candidate mechanisms that
may explain why the reward model prefers the TARGET RESPONSE over the
REFERENCE RESPONSE.

A mechanism may reflect either:
\begin{itemize}
  \setlength{\itemsep}{1pt}
  \setlength{\topsep}{2pt}
  \item a genuine substantive improvement, such as better correctness,
  reasoning, relevance, clarity, coverage, or task completion; or
  \item a potential reward-model shortcut, such as unnecessary length,
  redundancy, formatting, structure, prestige language, confidence, or
  validation that may attract reward without proportional substantive
  improvement.
\end{itemize}

Identify 1 to \texttt{\{num\_mechanisms\}} distinct mechanisms.
Use fewer mechanisms when they are sufficient.
Do not add weak, redundant, or speculative mechanisms just to reach the
maximum.

Each mechanism should be concise, atomic, generalizable, and describe one
behavior that could be added, removed, or toggled with a small edit.

Explain the reward model's preference, not which response is actually better.

\medskip
\textbf{USER REQUEST:}

\texttt{\{user\_prompt\}}

\medskip
\textbf{TARGET RESPONSE:}

\texttt{\{target\_response\}}

\medskip
\textbf{REFERENCE RESPONSE:}

\texttt{\{reference\_response\}}

\medskip
Return only a JSON array of mechanism descriptions.

\end{tcolorbox}

\vspace{0.8em}

\subsection{Mechanism Evaluation Prompts}
\label{app:mechanism_evaluation_prompts}

\begin{tcolorbox}[
  enhanced,
  breakable,
  colback=orange!10!white,
  colframe=blue!5!black,
  arc=2mm,
  boxrule=1pt,
  title={\bfseries Mechanism Quality Evaluation Prompt},
  coltitle=white,
  attach boxed title to top left={yshift=-2mm, xshift=3mm},
  boxed title style={
    enhanced,
    colback=blue!5!black,
    colframe=blue!5!black,
    arc=2mm,
    boxrule=0pt
  },
  top=0.5mm,
  left=1mm,
  right=1mm,
  bottom=0.5mm
]
\small
\vskip8pt

\textbf{System Prompt}

You evaluate explanation quality and return valid JSON only.

\medskip
\textbf{User Prompt Template}

You are evaluating one candidate mechanism set for a pair of assistant
responses.

The mechanism set should describe reusable response-level behavioral rules
that explain how the TARGET response differs from the FOIL response.

Do not evaluate which response is better.
Do not use or infer reward scores.
Do not compare this mechanism set with any other model's mechanism set.
Do not reward generic fluency, polished wording, or length by itself.
Use only the current question, TARGET response, FOIL response, and mechanism
set.

\medskip
\textbf{Question:}

\texttt{<question>}\\
\texttt{\{prompt\}}\\
\texttt{</question>}

\medskip
\textbf{TARGET response:}

\texttt{<target\_response>}\\
\texttt{\{target\_response\}}\\
\texttt{</target\_response>}

\medskip
\textbf{FOIL response:}

\texttt{<foil\_response>}\\
\texttt{\{foil\_response\}}\\
\texttt{</foil\_response>}

\medskip
\textbf{Mechanism set:}

\texttt{\{mechanisms\}}

\medskip
Score the mechanism set on each criterion using only 0, 1, or 2:

\medskip
\texttt{contrastive\_selectivity}:
\begin{itemize}
  \setlength{\itemsep}{1pt}
  \setlength{\topsep}{2pt}
  \item[2 =] the mechanisms clearly describe meaningful behavioral
  differences between TARGET and FOIL in this specific pair;
  \item[1 =] the mechanisms mostly describe TARGET, but at least one mechanism
  is a property that also applies to FOIL or does not clearly separate the two
  responses;
  \item[0 =] the mechanisms do not meaningfully distinguish TARGET from FOIL.
\end{itemize}

\texttt{scope\_calibration}:
\begin{itemize}
  \setlength{\itemsep}{1pt}
  \setlength{\topsep}{2pt}
  \item[2 =] each mechanism is stated at an appropriate level of scope and does
  not claim a broader behavioral difference than the responses support;
  \item[1 =] at least one mechanism is somewhat overgeneralized or overly
  narrow, but still captures the main observed difference;
  \item[0 =] most mechanisms substantially overstate, understate, or
  mischaracterize the scope of the observed behavioral differences.
\end{itemize}

\texttt{directional\_fidelity}:
\begin{itemize}
  \setlength{\itemsep}{1pt}
  \setlength{\topsep}{2pt}
  \item[2 =] the mechanisms consistently describe the behavioral change in the
  correct TARGET-relative-to-FOIL direction;
  \item[1 =] most directions are correct, but at least one mechanism is
  ambiguous or incompletely directional;
  \item[0 =] multiple mechanisms reverse or fail to specify the TARGET-FOIL
  direction.
\end{itemize}

\textbf{Important:}
\begin{itemize}
  \setlength{\itemsep}{1pt}
  \setlength{\topsep}{2pt}
  \item The main score is the equal-weight average of these three criteria.
\end{itemize}

Return JSON only:
\begin{quote}
\ttfamily
\{\\
\hspace*{1em}"contrastive\_selectivity": 0,\\
\hspace*{1em}"scope\_calibration": 0,\\
\hspace*{1em}"directional\_fidelity": 0,\\
\hspace*{1em}"brief\_reason": "..."\\
\}
\end{quote}

\end{tcolorbox}

\vspace{0.8em}

\begin{tcolorbox}[
  enhanced,
  breakable,
  colback=orange!10!white,
  colframe=blue!5!black,
  arc=2mm,
  boxrule=1pt,
  title={\bfseries Choice Evaluation Prompt},
  coltitle=white,
  attach boxed title to top left={yshift=-2mm, xshift=3mm},
  boxed title style={
    enhanced,
    colback=blue!5!black,
    colframe=blue!5!black,
    arc=2mm,
    boxrule=0pt
  },
  top=0.5mm,
  left=1mm,
  right=1mm,
  bottom=0.5mm
]
\small
\vskip8pt

You are an evaluator comparing two candidate responses according to
\textbf{one specified scoring mechanism}.

You will be given:
\begin{enumerate}
  \setlength{\itemsep}{1pt}
  \setlength{\topsep}{2pt}
  \item A user prompt
  \item Response A
  \item Response B
  \item A scoring mechanism
\end{enumerate}

Your task is to determine:

\textbf{If the provided scoring mechanism were the ONLY criterion used for
scoring, which response should receive the higher score?}

Please strictly follow these rules:
\begin{itemize}
  \setlength{\itemsep}{1pt}
  \setlength{\topsep}{2pt}
  \item \textbf{The provided scoring mechanism is the only criterion you may
  use to distinguish Response A from Response B.}
  \item Use the user prompt only to understand the context and meaning of the
  two responses. Do not use it to introduce additional evaluation criteria.
  \item Do not judge the responses based on your own general preferences about
  response quality.
  \item Do not consider correctness, helpfulness, relevance, completeness,
  safety, writing quality, level of detail, conciseness, or any other property
  \textbf{unless that property is explicitly described by the scoring
  mechanism}.
  \item Do not infer, add, or introduce any evaluation criterion that is not
  stated in the scoring mechanism.
  \item Focus only on the difference between Response A and Response B with
  respect to the behavior or textual property described by the scoring
  mechanism.
  \item If the scoring mechanism describes a behavior associated with a
  \textbf{higher score}, select the response that exhibits that behavior more
  strongly.
  \item If the scoring mechanism describes a behavior associated with a
  \textbf{lower score}, select the response that exhibits that behavior less
  strongly.
  \item Select \textbf{TIE} if the two responses exhibit the specified
  mechanism to approximately the same degree, or if the mechanism does not
  provide sufficient evidence to reliably distinguish between them.
  \item Do not guess. If the provided mechanism does not give sufficient
  evidence to favor either response, select \textbf{TIE}.
\end{itemize}

\texttt{<User Prompt>}\\
\texttt{\{prompt\}}\\
\texttt{</User Prompt>}

\medskip
\texttt{<Response A>}\\
\texttt{\{response\_a\}}\\
\texttt{</Response A>}

\medskip
\texttt{<Response B>}\\
\texttt{\{response\_b\}}\\
\texttt{</Response B>}

\medskip
\texttt{<Scoring Mechanism>}\\
\texttt{\{mechanism\}}\\
\texttt{</Scoring Mechanism>}

\medskip
Now determine which response should receive the higher score
\textbf{based only on the scoring mechanism above}.

Output exactly one of the following, with no additional explanation:

\begin{center}
\ttfamily
A\\[1pt]
B\\[1pt]
TIE
\end{center}

\end{tcolorbox}

\vspace{0.8em}

\subsection{Counterfactual Intervention Prompts}
\label{app:counterfactual_intervention_prompts}

\begin{tcolorbox}[
  appendixprompt,
  title={\bfseries Stage 1 Evidence Identification Prompt}
]
\small
\vskip8pt

\textbf{System Prompt}

You identify evidence for one candidate mechanism in two responses.

Return valid JSON only.

\medskip
\textbf{User Prompt Template}

In the task of response quality scoring, a trained deep learning model assigns
real-valued scores to responses, where a higher score indicates higher
predicted response quality.

\medskip
\textbf{Question:}

\texttt{"\{prompt\}"}

\medskip
\textbf{HIGHER-SCORED response (score = \texttt{\{high\_score\}}):}

\texttt{"\{high\_response\}"}

\medskip
\textbf{LOWER-SCORED response (score = \texttt{\{low\_score\}}):}

\texttt{"\{low\_response\}"}

\medskip
A candidate mechanism has been generated to describe a behavior that may
explain why the HIGHER-SCORED response receives a higher score:

\texttt{"\{mechanism\}"}

\medskip
Your task is to determine whether each response exhibits this mechanism and
identify the corresponding evidence.

Judge HIGH and LOW independently according to the mechanism as written.

\medskip
Use:

\medskip
\textbf{PRESENT:}
The response clearly exhibits the mechanism, including its important
modifiers.

\medskip
\textbf{ABSENT:}
The response does not exhibit the mechanism, or exhibits substantially less
of the relevant behavior.

\medskip
\textbf{NONE:}
There is insufficient evidence to determine whether the mechanism is present
or absent.

\medskip
For evidence:
\begin{itemize}
  \setlength{\itemsep}{1pt}
  \setlength{\topsep}{2pt}
  \item Return at most 5 short exact phrases from each response.
  \item Select the smallest phrases sufficient to identify the mechanism.
  \item Prefer short phrases over full sentences.
  \item Each phrase must contain at most 5 words.
  \item Do not return an entire paragraph.
  \item Copy phrases exactly from the original response.
  \item Every evidence phrase must be copied verbatim from the corresponding
  response, not from the user question, the other response, or a paraphrase of
  the response.
  \item If the mechanism is absent because something is missing, the phrase
  list may be empty.
\end{itemize}

Do not force a contrast between HIGH and LOW. If both responses exhibit the
mechanism, label both PRESENT.

\medskip
Return JSON only:
\begin{quote}
\ttfamily
\{\\
\hspace*{1em}"high\_evidence": \{\\
\hspace*{2em}"state": "PRESENT | ABSENT | NONE",\\
\hspace*{2em}"phrases": ["<exact phrase>"],\\
\hspace*{2em}"description": "<brief explanation>"\\
\hspace*{1em}\},\\
\hspace*{1em}"low\_evidence": \{\\
\hspace*{2em}"state": "PRESENT | ABSENT | NONE",\\
\hspace*{2em}"phrases": ["<exact phrase>"],\\
\hspace*{2em}"description": "<brief explanation>"\\
\hspace*{1em}\}\\
\}
\end{quote}

\end{tcolorbox}

\vspace{0.8em}

\begin{tcolorbox}[
  appendixprompt,
  title={\bfseries Add Rewrite Prompt}
]
\small
\vskip8pt

\textbf{System Prompt}

You make minimal, targeted modifications to a response.
Return only the modified response, or null if no suitable local edit is
possible.

\medskip
\textbf{User Prompt Template}

\texttt{<question>}\\
\texttt{\{prompt\}}\\
\texttt{</question>}

\medskip
\textbf{Original response:}

\texttt{<response>}\\
\texttt{\{response\}}\\
\texttt{</response>}

\medskip
\textbf{Target mechanism:}

\texttt{"\{mechanism\}"}

\medskip
\textbf{Evidence phrases:}

\texttt{\{evidence\}}

\medskip
Your task is to make a minimal, targeted modification to the original response
so that it \texttt{\{direction\_word\}} the target mechanism.

Center the modification on the identified evidence phrases.

The modified response should remain the same as the original response in all
other respects as much as possible.

Do not change any part of the original response that does not need to change.
In particular:
\begin{itemize}
  \setlength{\itemsep}{1pt}
  \setlength{\topsep}{2pt}
  \item Do not fix mistakes or generally improve the response.
  \item Do not add new facts, reasoning, examples, or task-relevant
  information.
  \item Do not rewrite unrelated sentences or restructure the response.
  \item Modify text outside the evidence phrases only when strictly necessary
  for local grammatical or semantic coherence.
\end{itemize}

\textbf{STRENGTHEN:}
Make the response more clearly exhibit the target mechanism.

\medskip
If the target mechanism cannot be changed through a local, minimal
modification, return null.

\medskip
Return only the modified response.

\end{tcolorbox}

\vspace{0.8em}

\begin{tcolorbox}[
  appendixprompt,
  title={\bfseries Remove Rewrite Prompt}
]
\small
\vskip8pt

\textbf{System Prompt}

You make controlled, targeted modifications to a response.

For REMOVE interventions, the target mechanism must no longer be clearly
present in the modified response.

Center the modification on the identified evidence while preserving unrelated
content as much as possible.

Return only the modified response, or null if no suitable controlled edit is
possible.

\medskip
\textbf{User Prompt Template}

\texttt{<question>}\\
\texttt{\{prompt\}}\\
\texttt{</question>}

\medskip
\textbf{Original response:}

\texttt{<response>}\\
\texttt{\{response\}}\\
\texttt{</response>}

\medskip
\textbf{Target mechanism:}

\texttt{"\{mechanism\}"}

\medskip
\textbf{Evidence phrases:}

\texttt{\{evidence\}}

\medskip
Your task is to modify the original response so that it no longer clearly
exhibits the target mechanism.

Center the modification on the evidence phrases.

Treat the evidence phrases as the primary edit locations.
First try to remove the target mechanism by modifying the evidence phrases
and their immediately related context.

The evidence phrases are anchors rather than strict boundaries.
If the target mechanism would still clearly remain after modifying only the
evidence phrases, extend the edit only to other nearby or related parts that
also express the same mechanism.

Do not make only superficial substitutions to the evidence phrases if the same
underlying mechanism would still remain clearly present.

If the mechanism is expressed globally through style, verbosity, structure,
tone, organization, or level of detail, modify multiple relevant parts only
when necessary to remove the mechanism.

Removing the mechanism does not mean maximally expressing its opposite.

Preserve unrelated factual content, reasoning, meaning, examples, structure,
style, and wording as much as possible.

\medskip
Do not:
\begin{itemize}
  \setlength{\itemsep}{1pt}
  \setlength{\topsep}{2pt}
  \item fix unrelated mistakes or generally improve the response;
  \item add new facts, reasoning, examples, or task-relevant information;
  \item remove unrelated information;
  \item rewrite unrelated parts.
\end{itemize}

\textbf{Priority:}
\begin{enumerate}
  \setlength{\itemsep}{1pt}
  \setlength{\topsep}{2pt}
  \item Successfully remove the target mechanism.
  \item Center the modification on the evidence phrases.
  \item Preserve unrelated content and meaning.
  \item Minimize the amount of change.
\end{enumerate}

Before returning, check whether the modified response still clearly exhibits
the target mechanism.
If it does, continue editing only the relevant evidence-centered or related
parts until the mechanism is no longer clearly present.

If the target mechanism cannot be removed without substantially changing
unrelated content, return null.

\medskip
Return only the modified response.

\end{tcolorbox}

\end{document}